\documentclass[preprints,article,accept,moreauthors,pdftex]{Definitions/class}

\usepackage{amsmath} 
\usepackage{amssymb}
\usepackage[utf8]{inputenc}
\usepackage[english]{babel}
\usepackage{graphicx}
\usepackage{adjustbox}
\usepackage{array}

\setitemize{parsep=6pt,itemsep=0pt,leftmargin=*,labelsep=5.5mm}
\setenumerate{parsep=6pt,itemsep=0pt,leftmargin=*,labelsep=5.5mm}
\setlist[description]{itemsep=0mm}

\usepackage{xcolor}
\usepackage{longtable}
\usepackage{tikz}
\usetikzlibrary{matrix, positioning}
\usepackage{soul}
\usepackage{multirow}

\usepackage{pdflscape}
\usepackage{afterpage}
\usepackage{caption}

\usepackage{booktabs}
\usepackage{pifont}
\usepackage{lscape}

\usepackage{upgreek}
\usepackage{textcomp}

\usepackage{hyperref}

\usepackage{booktabs}

\usepackage{tabularx}

\usepackage{amssymb}
\usepackage{latexsym}

\usepackage{url}
\usepackage{setspace}

\newcommand{\mscrev}[1]{{\color{black}{#1}}}

\firstpage{1} 
\pubvolume{1}
\issuenum{1}
\articlenumber{5}
\pubyear{2020}
\copyrightyear{2020}
\history{}
\updates{yes} 

\Title{
	When Deep Learning Meets Data Alignment: A Review on Deep Registration Networks (DRNs)
}

\Author{{Victor Villena-Martinez} $^{1,}$*, 
	Sergiu Oprea $^{1}$,
	Marcelo Saval-Calvo $^{1,}$*,
	Jorge Azorin-Lopez $^{1}$,
	Andres Fuster-Guillo $^{1}$
	and Robert B.   Fisher $^{2}$}

\AuthorNames{Victor Villena-Martinez, Sergiu Oprea, Marcelo Saval-Calvo, Jorge Azorin-Lopez, Andres Fuster-Guillo and Robert B.   Fisher}

\address{%
$^{1}$ \quad Department of Computer Technology, University of Alicante, 03690 Alicante, Spain; {soprea@dtic.ua.es (S.O.); jazorin@dtic.ua.es (J.A.L.); fuster@dtic.ua.es (A.F.G.)}\\ 
$^{2}$ \quad School of Informatics, University of Edinburgh, Edinburgh EH8 9AB, UK; {rbf@inf.ed.ac.uk}}

\corres{Correspondence: vvillena@dtic.ua.es (V.V.M.); msaval@dtic.ua.es (M.S.C.)}

\abstract{This paper reviews recent deep learning-based registration methods.   Registration is the process that computes the transformation that aligns     datasets, and the accuracy of the result depends on multiple factors.   {The most significant factors are the size of input data; the presence of noise, outliers and occlusions; the quality of the extracted features; real-time requirements; and the type of transformation, especially those defined by multiple parameters, such    as non-rigid deformations.} Deep Registration Networks (DRNs) are those architectures trying to solve the alignment task using a learning algorithm.   In this review, we classify these methods according to a proposed framework based on the traditional registration pipeline.   This pipeline consists {of four steps: target selection, feature extraction, feature matching, and transform computation for the alignment.}{ This new paradigm introduces a higher-level understanding of registration, which makes explicit the challenging problems of traditional approaches.   The main contribution} of this work is to provide a comprehensive starting point to address registration problems from a learning-based perspective and to understand the new range of possibilities.
}


\begin{document}


%



\section{Introduction}
\label{sec:introduction}

In the context of computer vision, registration is the process of aligning data into a common frame of reference.   In other words, it aligns     datasets---captured from different sources, viewpoints, and/or at a different time step---by means of geometric transformations.   Here,  we consider both 2D and 3D data, and data in point sets, grids, and meshes.   Rigid and non-rigid registration has already been widely addressed in the computer vision literature through potential applications mostly for data analysis such as body modeling \cite{VillenaMartinez2017} for pose analysis; computed tomography registration \cite{Zeman1994,Boldea2008} for medical diagnosis; multi-camera registration for robot guidance \cite{CuevasVelasquez2018}; and applications in object classification on assembly lines \cite{Zhao2019}, among others.   In the aforementioned applications, registration represents a crucial component.   It fuses a vast amount of raw data captured under different scenarios, which greatly facilitates the analysis process.

The growing number of available consumer-grade devices, such as RGB-D cameras and LiDAR sensors, provides  quasi-unlimited and cheap data of different modalities.   However, the raw data must be previously structured, either hierarchically or semantically, to extract high-level information.   The vast amount of data has surpassed the potential of the traditional registration paradigm and led researchers to consider learning-based approaches.   Dealing with a huge amount of raw and unstructured multidimensional data is not straightforward, yet they satisfy Deep Learning (DL) methods that are known to be data-hungry.   Learning-based approaches have proliferated in recent years, making great strides {in different fields} \cite{Garcia-Garcia2018,oprea2020review,lu2020deep}.   Considering this success, deep learning-based registration approaches are poised to leap over the previous state of the art in the registration paradigm.   However, existing DL-based techniques for rigid and non-rigid registration, mostly in an n-dimensional space, are far from accurate and reliable.   Furthermore, the direct application of DL techniques to the problem of registration is not straightforward; its lack of maturity and the rapid state of this field make  it difficult to keep up with the latest trends and track them properly.   

\subsection{Review Scope}

This paper reviews state-of-the-art learning-based approaches to registration.   The ability of deep neural networks to generalize from training data and manage geometric properties has created a new subfield  at the intersection between learning and registration algorithms.   Although some reviews of registration have been performed \cite{Tam2013,Zhu2019,Salvi2007}, no reviews   address  learning-based approaches for registration without focusing on a specific scope such    as medical image registration or image localization.

{The contributions of this paper are: (1) we provide a global overview of learning-based registration methods by proposing a well-defined framework that encompasses both the traditional and learning-based approaches; and (2) we review the recent learning-based registration approaches, which have been classified according to a proposed taxonomy to foster discussion.}



\begin{figure}[!t]
	\centering
	\includegraphics[width=0.8\textwidth]{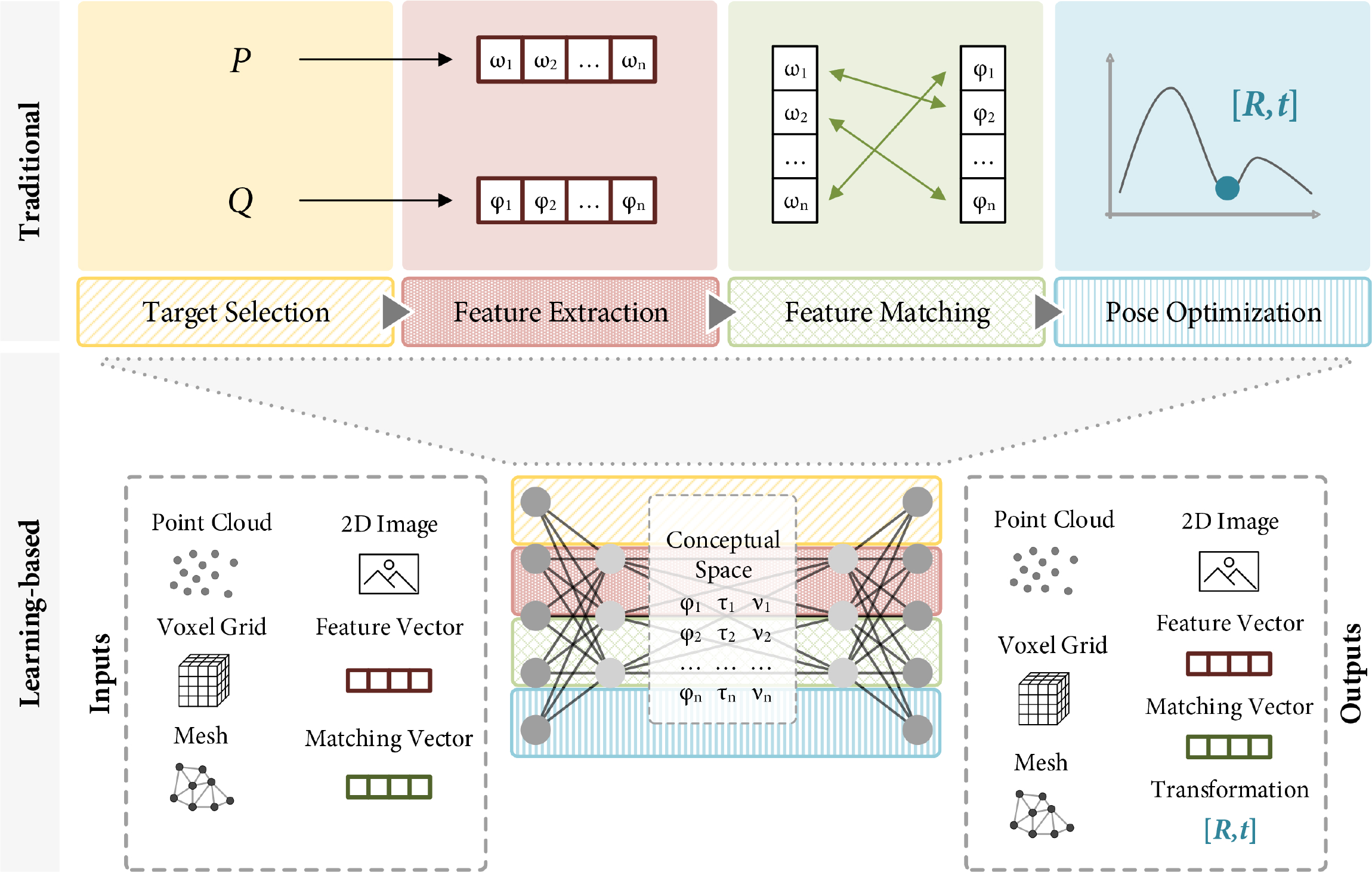}
	\caption{Registration framework.   Schematic of the registration process in the traditional pipeline on the top part, and the learning-based approaches reviewed in this paper at the bottom.   Learning-based approaches encapsulate the traditional pipeline in a conceptual space, allowing different types of inputs depending on the data and the number of encapsulated steps.   The conceptual space is the learned parameters during the training process, and theoretically could also be considered as an input for the registration process.}
	\label{fig1} 
\end{figure}

{
Figure \ref{fig1} graphically summarizes the scope of this paper.   The figure shows at the top the four main stages of the traditional pipeline for registering two given inputs ($P$ and $Q$).   These stages are: Target Selection (yellow, oblique lines), which defines the fixed input that the other input is going to be aligned to; Feature Extraction (red, dotted pattern),  which computes the set of features $\upomega$ and $\upvarphi$ for each input; Feature Matching (green, squared lines) to find correspondences between the previously extracted features; and Pose Optimization (blue, vertical lines), which is the process to minimize the distance error between both inputs.   The right end is the final result transformation ($[R,t]$), which is the rotation $R$ and translation $t$ parameters that indicate how data should be transformed to be aligned.   These stages and terminology are further explained in Section \ref{sec:registration_framework}.}

Vertically, the schematic is divided at the top with the traditional stages clearly defined and at the bottom the learning-based approaches that are reviewed here, represented by a neural network, and the possible data for inputs and outputs.   Having in mind this information, the inputs for the learning-based proposal could be in different formats, such    as point clouds, voxel grids, meshes, etc.      In addition to    full end-to-end approaches, some methods accept as inputs the result of the Feature Extraction or Matching stages in the traditional pipeline.   The output could be some data in a specific format as well as the result of the stages from the traditional pipeline (feature vector, matching vector, and transformation).
{Nevertheless, with the new learning-based approaches, a new conceptual kind of space appears, containing learned properties about the object, materials, and their behavior that can be registered with the input data (e.g., aligning input data of an inflated ball with its deflated state restricted by the physical behavior of the object, rather than aligning with a final deflated target).} The conceptual space is modeled by a neural network and its training process, and, theoretically, it could be considered as an input to the registration process,  but it is not an input that one might use in every registration instance since it is an internal representation.   This allows the network to encode conceptual models such    as physical phenomena (e.g., force vector and  symbolic/conceptual information such as ``sporty, comfy'') or mathematical rules.   Besides, the neural network could perform one or more phases from the traditional pipeline (represented by the colored rectangles (see Figure \ref{fig1}) shown in the network).   Considering all of this, we can see that there are multiple possibilities, combining inputs from different stages and outputs from the learning-based approaches.   

{
\subsection{Developments Relevant to Registration}
The number of works addressing registration with learning techniques has increased in recent years.   To identify them, strategic searches    were  performed in Scopus, Web of Science (WoS), and ArXiv.   The results obtained from Scopus and WoS come from indexed journals, which means that they have passed a peer-review process.   However, most of the recent works in this review   were reached through ArXiv, which is a preprint repository without peer review.   This is a double-edged sword.   ArXiv allows disseminating the work immediately while it is being reviewed for publication in a journal.   However, as some works here studied were not yet reviewed in another place, the authors had to perform a more in-depth evaluation.

Figure \ref{fig:evolution} shows the research papers published in the intersection of deep learning and registration involving 3D data over the last years in each repository.   It is noticeable how the number of publications has increased over the last three years.   Several search strings were employed to identify the methods surveyed in this paper.   The keywords grouped by the concept are:

\begin{itemize}
	\item To include learning-based methods: \textit{deep learning}; \textit{machine learning}.
	\item To indicate the data type employed: \textit{3D}; \textit{point cloud}; \textit{mesh}.
	\item To gather registration proposals: \textit{registration}; \textit{alignment}; \textit{transformation}; \textit{reconstruction}.
\end{itemize}

The search strings were designed by combining the words from the previous groups by choosing one from each.   

In the following sections, an analysis of learning-based approaches for registration is performed using a workflow extracted from traditional solutions.   We gather in Table \ref{tab:methods} the reviewed methods showing the individual properties for each one.   These approaches allow more complex inputs such    as conceptual models as well as 3D datasets.   However, since each proposal uses different datasets, a fair quantitative comparison cannot be done.

\newpage
\paperwidth=\pdfpageheight
\paperheight=\pdfpagewidth
\pdfpageheight=\paperheight
\pdfpagewidth=\paperwidth
\newgeometry{layoutwidth=297mm,layoutheight=210 mm, left=2.7cm,right=2.7cm,top=1.8cm,bottom=1.5cm, includehead,includefoot}
\fancyheadoffset[LO,RE]{0cm}
\fancyheadoffset[RO,LE]{0cm}

\newcommand{\ra}[1]{\renewcommand{\arraystretch}{#1}}
\renewcommand{\tabcolsep}{5pt}
\newcommand{\xmark}{\ding{55}}
\newcolumntype{C}[1]{>{\centering\arraybackslash}m{#1}}



\begin{table}[!t]
\centering

\tablesize{\tiny} 

\caption{\label{tab:methods}{Summary of the reviewed methods.   It shows the application, inputs, and outputs; the employed datasets; and the architecture of the network}.   The right columns record the stage of the traditional registration pipeline that the network addresses (\textbf{Ta}, \textbf{Ta}rget; \textbf{Fe}, \textbf{Fe}atures; \textbf{Ma}, \textbf{Ma}tching; \textbf{Tr}, \textbf{Tr}ansform).} 

\begin{tabular}{@{}lc>{\centering}m{0.15\linewidth}>{\centering}m{0.15\linewidth}>{\centering}m{0.15\linewidth}>{\centering}m{0.17\linewidth}cC{0.015\linewidth}C{0.015\linewidth}C{0.015\linewidth}C{0.015\linewidth}@{}}

 \toprule
 & & & \multicolumn{4}{c}{\textbf{Network Details}} & \multicolumn{4}{c}{\textbf{Pipeline}} \\
 \cmidrule(l){4-11}
 \textbf{Proposal} & \textbf{Year} & \textbf{Application} & \textbf{Inputs} & \textbf{Outputs} & \textbf{Datasets} & \textbf{Architecture \footnotemark} & \textbf{Ta} & \textbf{Fe} & \textbf{Ma} & \textbf{Tr} \\
 \midrule
	
	
	\citet{Yumer2016} & 2016 & Shape Deformation & Point Cloud / Label & Flow / Voxel Grid & ShapeNet \cite{chang2015shapenet}, SemEd \cite{Yumer2015} & CNN & \checkmark & \checkmark & \checkmark & \checkmark \\ 
	
	\citet{Elbaz2017} & 2017 & Descriptor & Depth Map & Reconstructed Depth Map & Challenging Datasets for Point Cloud Registration Algorithms \cite{Pomerleau2012} & AE & \xmark & \checkmark & \xmark & \xmark \\ 
	\citet{Li2017} & 2017 & MR Image Registration & (MR) Voxel Grid (x2) & Registered Voxel Grid & Alzheimer's Disease Neuroimaging Initiative (ADNI) \footnotemark & FCN & \xmark & \checkmark & \checkmark & \checkmark \\
	\citet{Wang2017} & 2017 & 3D Reconstruction from 2D Image & 2D Image & 3D Model (Voxel Grid) & ShapeNet \cite{chang2015shapenet}, PASCAL3D \cite{Xiang2014}, SHREC 13 \cite{li2013shrec} & GAN & \checkmark & \checkmark & \checkmark & \checkmark \\ 
	\citet{Zeng2017} & 2017 & Geometric Descriptor & Voxel Grid & Feature Vector & Analysis-by-Synthesis \cite{Valentin2016}, 7-Scenes \cite{Shotton2013}, SUN3D \cite{Xiao2013}, RGB-D Scenes v2 \cite{Lai2014}, \citet{Halber2017} & CNN & \xmark & \checkmark & \xmark & \xmark \\
 
	\citet{Ding2018} & 2018 & Multiple Point Clouds Registration (Localization) & Point Clouds & Discrete Occupancy Map & Active Vision Dataset \cite{Ammirato2017} & CNN & \checkmark & \checkmark & \checkmark & \checkmark \\ 
	\citet{Groueix2018} & 2018 & Matching Deformable Shapes & Point cloud & Point Cloud & SMPL \cite{Loper2015}, SURREAL \cite{Varol2017}, SMAL \cite{Zuffi2017}, FAUST \cite{Bogo2014}, TOSCA \cite{bronstein2008numerical}, SCAPE \cite{Anguelov2005} & SDN & \xmark & \checkmark & \checkmark & \checkmark \\ 
	\citet{Gundogdu2018} & 2018 & Garment PBS & Point Cloud / Mesh & Translation Vector & GarNet \footnotemark, SMPL \cite{Loper2015} & PointNet & \xmark & \checkmark & \checkmark & \checkmark \\ 
	\citet{Hanocka2018} & 2018 & Shape Alignment & Shapes (x2) & Transformed shape & ShapeNet \cite{chang2015shapenet}, COSEG \cite{Wang2012} & CNN & \xmark & \checkmark & \checkmark & \xmark \\ 
	\citet{Hermoza2018} & 2018 & 3D Reconstruction of Incomplete Objects & Voxel Grid (incomplete shape) / Label & Voxel Grid & ModelNet10 \cite{ZhirongWu2015}, 3D Pottery dataset \cite{Koutsoudis2010}, Custom Data & GAN & \checkmark & \checkmark & \checkmark & \checkmark \\ 
	\citet{Yew2018} & 2018 & Descriptor & Point Clouds & Local Descriptors & Oxford RobotCar \cite{Maddern2016}, KITTI Dataset \cite{Geiger2012}, ETH Dataset \cite{Pomerleau2012} & Siam.   CNN & \xmark & \checkmark & \checkmark & \xmark\\
	\citet{Kuang2018} & 2018 & 3D Medical Image Registration & Voxel Grid (x2) & Voxel Grid & MindBoggle101 \cite{Klein2012} & STN & \xmark & \checkmark & \checkmark & \checkmark \\ 
	\citet{Lin2018} & 2018 & Image Compositing & RGBA Foreground / RGB Background & 8 Dimensional Warp Parameter & CelebA \cite{Liu2015}, SUNCG \cite{Song2017} & ST-GAN & \xmark & \checkmark & \checkmark & \checkmark \\ 
	\citet{Litany2018} & 2018 & Body Shape Completion & Partial Mesh & Completed Mesh & DFAUST \cite{Bogo2017} & VAE & \checkmark & \checkmark & \checkmark & \checkmark \\ 
	\citet{Liu2018} & 2018 & Point Cloud Flow Estimation & Point cloud (x2) & Scene flow (point level) & FlyingThings3D \cite{Mayer2016} & CNN & \xmark & \checkmark & \checkmark & \xmark \\ 
	\citet{Mahapatra2018} & 2018 & Multimodal Image Registration & 2D medical multimodal images (x2) & Transformed Image & Retinal Images \cite{HajebMohammadAlipour2012}, Sunybrook \cite{radau2009evaluation} & GAN & \xmark & \checkmark & \checkmark & \checkmark \\ 
	\citet{Ofir2018} & 2018 & Multi-spectral 2D Descriptor & RGB / NIR & Pair of Features & CIFAR-10 \cite{krizhevsky2009learning}, \citet{Brown2011} & Siam.   CNN & \xmark & \checkmark & \checkmark & \xmark \\ 
	\citet{Yan2018} & 2018 & MR and TRUS Registration & MR Images (x2) & Transformation / Quality Check & Custom Data & GAN & \xmark & \checkmark & \checkmark & \checkmark \\ 
	\citet{Wang2018b} & 2018 & Force Simulation & Voxel Grid & Deformed 3D Model & Custom Data & VAE + AT & \checkmark & \checkmark & \checkmark & \checkmark \\ 

	\citet{Aoki2019} & 2019 & Point Cloud Registration & Point Clouds (x2) & Transformation & ModelNet40 \cite{ZhirongWu2015} & MLP + PointNet & \xmark & \checkmark & \checkmark & \checkmark \\ 
	\citet{Chang2019} & 2019 & Point Cloud Rigid Registration & Features & Transformation & Custom Data & CNN & \xmark & \xmark & \checkmark & \checkmark \\
 \citet{Guan2019} & 2019 & Vascular Image Registration & 3D CT / 2D DSA & 3D Transformation (translation and rotation) & Custom Data & MCNN & \xmark & \checkmark & \checkmark & \checkmark \\ 
	\citet{Jack2019} & 2019 & 3D Reconstruction from Single Image & 2D Image / Mesh & Mesh & ImageNet \cite{Deng2009}, ShapeNet \cite{chang2015shapenet} & CNN & \xmark & \checkmark & \checkmark & \checkmark \\ 
	
	\citet{Schaffert2019} & 2019 & Correspondence Weighting & Local Features & Weights Vector & Custom Data & CNN & \xmark & \xmark & \checkmark & \xmark \\ 
	\citet{Smirnov2019} & 2019 & 3D Reconstruction from 2D Sketch & 2D Shape & Mesh & ShapeNet \cite{chang2015shapenet} & CNN + MLPs & \checkmark & \checkmark & \checkmark & \checkmark \\ 
	\citet{Yang2019} & 2019 & Point Cloud Generation & Point Cloud & Point Cloud & ShapeNet \cite{chang2015shapenet} & AE & \checkmark & \checkmark & \checkmark & \checkmark \\ 
 \citet{Wang_2019_ICCV} & 2019 & Rigid Registration & Point Clouds (x2) & Transformation & ModelNet40 \cite{ZhirongWu2015} & CNN & \xmark & \checkmark & \checkmark & \checkmark \\ 
 \citet{wang20193dn} & 2019 & Deformation & 3D Mesh / 2D Image or Point Cloud & Mesh & ShapeNet \cite{chang2015shapenet} & PointNet + MLPs & \xmark & \checkmark & \checkmark & \checkmark \\
 \citet{wang2019coherent} & 2019 & Non-rigid Registration & Point Set (2D or 3D) (x2) & Aligned Point Set & Custom Data & MLPs & \xmark & \checkmark & \checkmark & \checkmark \\
 
 \citet{Pais2019} & 2020 & 3D Scan Registration & 3D Correspondences Vector & Weights Vector / Rotation and Translation & ICL-NUIM \cite{SungjoonChoi2015}, SUN3D \cite{zhou2014learning} & PointNet + ResNet & \xmark & \xmark & \checkmark & \checkmark \\ 
 
 \citet{li2020deterministic} & 2020 & Rigid Registration & Point Clouds (x2) & Transformation & ModelNet40 \cite{ZhirongWu2015} & PointNetLK & \xmark & \checkmark & \checkmark & \checkmark \\
	
 \citet{yuan2020deepgmr} & 2020 & Rigid Registration & Point Clouds & GMM Correspondences & ModelNet40 \cite{ZhirongWu2015}, Augmented ICL-NUIM \cite{SungjoonChoi2015,handa2014benchmark} & PointNet & \xmark & \checkmark & \checkmark & \xmark \\
 \citet{zhang2020deform} & 2020 & Multi-modal Deformable Registration & Voxelgrid (x2) & Transformation & Brain Tumor Segmentation (BraTS) \cite{menze2015multimodal} & GAN & \xmark & \checkmark & \checkmark & \checkmark \\


 \bottomrule
 
 \end{tabular}\\
 \begin{tabular}{@{}c@{}} 
\multicolumn{1}{p{\textwidth -.88in}}{\footnotesize $^1$ The architectures of some proposals are variants of the family identified in this column; $^2$ Alzheimer’s Disease Neuroimaging Initiative (ADNI) database (\url{http://adni.loni.usc.edu}); $^3$ GarNet dataset \url{https://cvlab.epfl.ch/research/garment-simulation/garnet/}.}
\end{tabular}
 
\end{table}

\newpage
\restoregeometry
\paperwidth=\pdfpageheight
\paperheight=\pdfpagewidth
\pdfpageheight=\paperheight
\pdfpagewidth=\paperwidth
\headwidth=\textwidth

\begin{figure}[!t]
	\centering
	\includegraphics[width=0.7\textwidth]{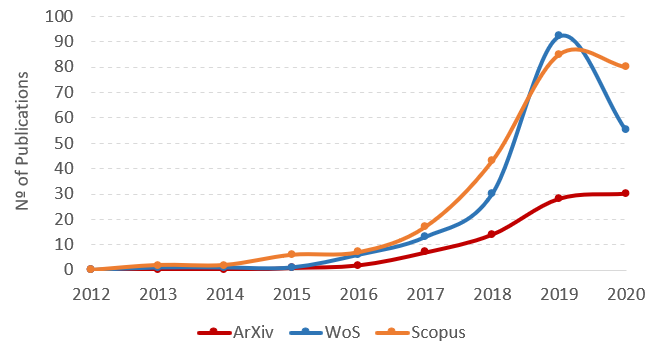}
	\caption{Research works published by year addressing registration of 3D data with learning approaches in each source   (data obtained through the search engines of each source through September 2020).}
	\label{fig:evolution} 
\end{figure}


}





\section{Registration Framework}
\label{sec:registration_framework}

\mscrev{
Registration aims to compute the alignment between     datasets.   Given two inputs \emph{P} = $\{\upomega_1, \upomega_2, \upomega_3,..., \upomega_n\}$ and $Q= \{\upvarphi_1, \upvarphi_2, \upvarphi_3,..., \upvarphi_m\}$, a registration process finds a transformation function $\chi$ that minimizes the alignment error between $P$ and $Q$, through checking the distance error between a pair of correspondences $(\upvarphi_i,\upomega_j)$ of each input, as shown in Equation (\ref{eq2}).   There are different error and distance functions ($dist$), e.g., perpendicular distance rather than Euclidean distance or Huber distance, $L_1$ error, etc.   
}



\begin{equation}
\label{eq2}
 E_P = \sum_{i,j}^{n,m} gate(\upvarphi_i,\upomega_j) * dist( \upvarphi_i - \chi(a,\upomega_j) )
\end{equation}
\mscrev{{, }Let $\chi$ transform each element of $P$, according to the transformation parameters $a$, with the goal of minimizing the error $E_P$ between $P$ and $Q$, while $gate(\upvarphi_i,\upomega_j) = 1$ if the features correspond and $0$ otherwise.   
}

In the literature, the terms registration and  reconstruction (or shape completion) are sometimes employed to refer to the same process.   The main difference is that reconstruction is at a higher level than registration since the registration process is a part of reconstruction methods, but a reconstruction method may not perform any registration.   {That is, registration aims to find the transformation to align data, while the main goal of reconstruction is to obtain a virtual representation from the scene.   To this end, a reconstruction method may not include a registration process, for example, those   that only use a single view to obtain the virtual representation.}
Although these definitions previously had clear differentiation, now learning-based approaches have blurred the line between them, e.g., in \cite{Litany2018}, an algorithm designed to perform shape completion is tested by performing registration tasks.

According to \citet{Tam2013}, the registration process can be divided into three core components: target selection, correspondences and constraints, and optimization.   This sequence has been used often by registration algorithms to find the alignment of 3D datasets.   
These stages are shown in Figure \ref{fig1}.   A more detailed classification was presented by \citet{SavalCalvo2015}, including pre-processing and post-processing phases:

\begin{itemize}
 \item Pre-processing.   This stage adapts the input data to meet the requirements of the algorithms.   
 
 
 
 \item Target Selection.   It is often necessary to differentiate between the dataset that will remain fixed and the one that will be moved towards the fixed set to perform the alignment. In the literature, different nomenclatures could be found for these {fixed/moving} terms such as model/data, anchor/moving,  or target/source.   
 
 
 
 \item Feature Extraction.   This stage refers to the process of finding those landmarks or salient features that will be used to calculate the matches between sets.
 
 \item Feature Matching.   It refers to the identification of corresponding features between the target and each moving data.   The pair composed of pairs of features is called a correspondence.
 
 \item Pose Optimization.   Here, the algorithm computes the transformation that minimizes a distance between the correspondences, aligning the sets into a common reference space.   
 
 \item Post-processing.   This step is highly dependant on the problem itself.   It could include global optimization, such as loop-closure, data-cleaning in solid mesh estimation, surface extraction, or outlier removal.
\end{itemize}

\section{Deep Learning in the Context of Registration}
\label{sec:dl_in_registration}

Deep Learning is the subfield of Machine Learning that studies Deep Neural Networks (DNN), which increases the number of hidden layers in a Neural Network (and potential layer-to-layer transformations) and computes multiple levels of abstraction.   It transforms the data in a non-linear fashion by learning complex functions and transformations \cite{LeCun2015}.
An extended review of the history of deep learning and its approaches can be found in \cite{Alom2018}.

 
In a similar way to humans, neural networks are able to \textit{understand} the input data by extracting an abstract understanding of it \cite{LeCun2015}.   
 \citet{Bench-Capon1990} considered the \textit{{representation of knowledge}} remaining in a learning system after the training process with the following definition: \textit{{a set of syntactic and semantic conventions that makes it possible to describe things.}} This representation of knowledge could be understood as a conceptual model of the object.   \citet{norman1983} defined conceptual models as an accurate, consistent, and complete representations of knowledge, coherent with the real world and physics rules.   There is a gap between an observed phenomenon and the mathematical model.   According to \citet{nersessian1992scientists},  mental models are located in this gap, but they can be incomplete or unscientific.   A mathematical model is also a conceptual model, which is an external representation that facilitates the comprehension of a teaching system.   It is functional and coherent with scientific knowledge \cite{Greca2000}.

 This conceptual knowledge can improve alignment problems.   Traditional registration approaches have different challenges, that lead into one general limitation: the lack of generalization of these algorithms.   Usually they are highly dependent on the correspondences between the input datasets.   
With the development of DL, the \textit{{remaining knowledge}}, defined before as a set of syntactic and semantic conventions, could be considered as a conceptual model, that, in the case of registration processes, could be a target to align with spatial data.   Theoretically, the idea of the conceptual model allows to differentiate the input data of a registration process into defined or non-defined models.   The defined aspects are models that represent specific spatial data (commonly 2D or 3D) while a non-defined model is a generalization of a dataset produced by a learning system, e.g., the concept of a ball, those properties that make an object a ball, rather than the specific instance of a ball itself using geometrical aspects.

\begin{figure}[H]
	\centering
	\includegraphics[width=0.6\textwidth]{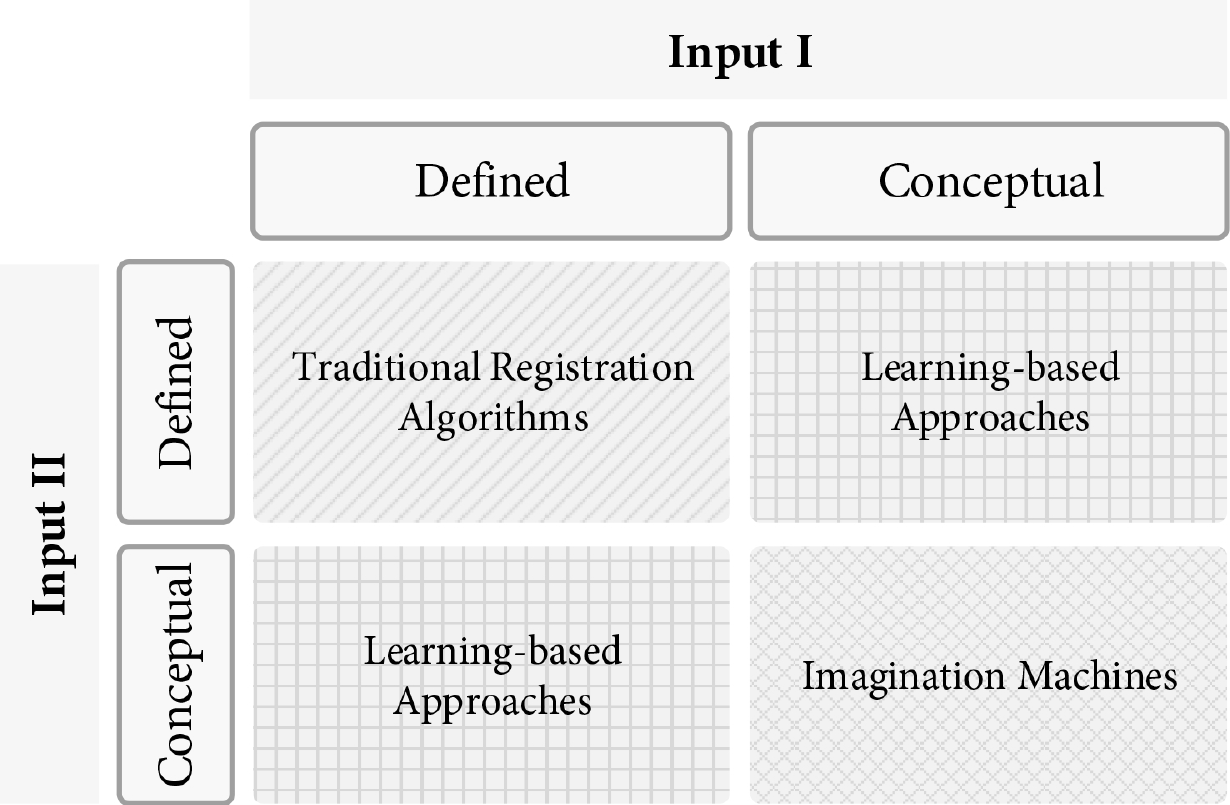}
	\caption{The taxonomy present in registration algorithms as a result of the intersection between defined and conceptual/non-defined features.}
	\label{tab:taxonomy} 
\end{figure}

The conceptual models have also been applied in registration, for example, in the work of \citet{Yumer2016}, in which the network learns properties of objects, being able to know what a more sporty car looks like or a more comfortable chair is, and modifying a 3D model to fit those properties while preserving the main features of the original data.   
With this approach, three combinations of input information are possible: \textit{defined model}/\textit{defined model}, \textit{defined model}/\textit{conceptual model}, and \textit{conceptual model}/\textit{conceptual model}.   This taxonomy is shown in Figure \ref{tab:taxonomy}.
The classical algorithms for registration are included in the first of the possibilities, one input is used as a target or as a reference set whilst the other is transformed to be aligned with the first, but always with defined data.   By contrast, the use of neural networks for registration results in other combinations where conceptual models are included.
Those models need not be specifically defined, e.g., they can be synthesized by a trained network with the learned features coming from the training data.   Then, these features can be used in the registration process afterward or even in the same network.   In any case, there is no need to understand the working space of the network.   Its internal representation is alien to human knowledge.

The combination of two conceptual models could be possible with the growth of \textit{Imagination Machines} proposed by \citet{mahadevan2018imagination}, which aims to provide artificial intelligence systems flexibility and connections between the learned aspects through training processes not based on labels and classifications of the input data.


\section{Review of Learning-Based Approaches for Registration}
\label{sec:review}

To show the advancements of neural networks applied to registration problems, an analysis of recent works in the intersection of these fields is performed.   To establish a comparative framework between these new approaches, we use the workflow abstracted from traditional solutions, as introduced in Section \ref{sec:introduction}.   This workflow is divided into four stages: target, features, matching, and optimization.   In this section, an analysis of learning approaches that use some or all of these stages of the workflow is performed.

The reviewed methods are shown in Table \ref{tab:methods}, classified regarding the traditional phases of a registration workflow.   The first columns are the application, inputs, outputs,   employed datasets, and   architecture of each method.   The final columns indicate the parts of the traditional registration process implemented in the DN.   In this way, the target column refers to the need for the method to have anchor data as a target to perform the alignment.   If there is no checkmark in a column, it means the method requires a defined target as input, otherwise the target is a conceptual model inside the network.   
This is possible if the generalization of the training data is implicit in the network knowledge, i.e., the main properties from the inputs are learned by the network.   For example, it is possible to think of a specific instance of a chair or think of a chair as a concept, with those properties that a chair must fulfill to be considered as such.

The feature column indicates the ability of the method to find features in the data using a neural network, such    as the work of \citet{Ofir2018}.   The next step in the workflow is the matching between features.   There are some proposals which train a network to be able to check the accuracy of the correspondences; \citet{Pais2019} proposed  a network to align two     datasets given the features and the matching between them.   The network determines if the features are correct, removing some of them if necessary.   The last column indicates the ability of the learning approach managing the geometric operations that align     datasets, such    as computing the camera pose \cite{Ding2018} or the transformation parameters \cite{Gundogdu2018}.

In the recent state-of-the-art methods, we find methods that carry out the whole registration, i.e., they cover the main parts of the traditional pipeline of registration, as well as other methods that implement only some stages of the pipeline.   This classification of the analyzed works has been done as a way to compare them in a common framework using a traditional perspective of the registration methods, going into detail in the key aspects of each method according to the stage where it contributes a novel method.


\subsection{Target Level}
At the target level, there are methods that generalize from the training process, exploring the idea of the conceptual model.   This enables registration using learning approaches.   For instance, the work of \citet{Yumer2016} {uses  semantically deforming shapes in 3D through free-form deformation using lattices}.   It does not use a target model to deform a mesh, the properties of the target are learned by the proposed network.   The network is able to perform non-rigid deformations over 3D models to fit a given semantic property.   As a result, it provides the deformed 3D model as well as the deformation flow of the data to fit that model while preserving original details.   Similarly, a key aspect of the unsupervised learning approach proposed by \citet{Ding2018} is that there is no target for the registration process.   The network is capable of locating each input point cloud in a global space, solving SLAM problems in which multiple point clouds have to be registered rigidly.   {The employed architecture is discussed in the following sections.}

Adversarial training (AT), Autoencoders (AE), and Generative Adversarial Networks (GANs) are used in some works for extracting conceptual descriptions.   For instance, \citet{Wang2017} employed  an {adversarial approach with CNN} networks to reconstruct a 3D model of an object from its 2D image.   {The key aspect of this work is the combination of a 2D autoencoder-based network with a deconvolutional network.   The first network transforms the input image to the latent space, while the second transforms from the latent space to 3D space, acting as a 3D generator.   It is an unsupervised generative neural network that accurately predicts 3D volumetric objects from single real-world 2D images.} The network has learned multiple objects and internally performs the registration between the image and the conceptual model.   In a similar way, \citet{Hermoza2018} also used a GAN network for predicting the missing geometry of damaged archaeological objects, indicating the reconstructed object in a voxel grid format and a label designating its class.   {Its network architecture combines a completion loss and an improved Wasserstein GAN loss.}

{\citet{Smirnov2019} proposed  a method to generate a 3D model from a 2D sketch.   The 3D models are defined by a set of parametric patches.   They employed an encoder-style architecture using convolutional layers and residual blocks that generates a series of 3D patches from the sketch, then a set of MLPs carry out the intersection between patches.   In this work, registration between different spaces is performed with the provided sketch and the internal knowledge that comes from the training procedure.} Similarly, the generative model of \citet{Yang2019} uses a variation of an autoencoder architecture to generate 3D point clouds by modeling them as a distribution of distributions.   Concretely, their method learns the distribution of shapes at the first level, and the distribution of points given a shape at the second level.   As a result, the method is able to generate points as a given shape by parameterizing the transformation of points from an initial Gaussian distribution of them.   Moreover, Variational Autoencoders (VAEs) are being used in an adversarial training framework, such    as the work of \citet{Wang2018b}.   In this case, they are employed for predicting structural deformations produced by forces given a single depth image and the conditions of the input, which includes properties of the material, the strength of the force, its location, etc.   {The generator predicts the force over a 3D model, and the discriminator, used for training, should determine if the applied force comes from the generator or from the ground-truth}.   This approach enables the network to learn non-rigid deformations and it can generalize the deformations to {unknown objects having into account properties of the materials}.   Other approaches are able to train a variational autoencoder with graph convolutional operations for completing missing data from partial body shapes while dealing with non-rigid deformations \cite{Litany2018}.   {They are able to identify the output space of the generator that best aligns with the partial input.   Partial shapes are completed by deforming a randomly generated shape to be aligned with a partial input.   This approach is robust to non-rigid deformations and has the ability to reconstruct missing data.   It shows topology understanding by the encoder--decoder architecture.}


\subsection{Feature and Matching Level}


Learning approaches have demonstrated successful results in performing feature extraction and matching for registration purposes.   Auto-encoders have been used for feature extraction.   For instance, \citet{Elbaz2017} used  in their proposal for point cloud registration a Deep Auto Encoder (DAE) for extracting low-dimensional descriptors from large scale point clouds.   {The training of the DAE is unsupervised, and it} is able to extract a compact representation from depth maps that capture the significant geometric properties of the input data.{ \citet{yuan2020deepgmr} proposed  the method DeepGMR to perform the registration by matching points to a probability distribution whose parameters are estimated by a neural network from the input point clouds.   That network learns latent correspondences between points and Gaussian Mixture Model (GMM) components that are pose-invariant.   The network estimates point-to-component correspondences, following two compute blocks to obtain the GMM parameters and the transformation.} \citet{Groueix2018} introduced Shape Deformation Networks (SDNs) {in an encoder--decoder architecture for matching deformable shapes}, where the encoder is able to extract a global shape descriptor from a 3D model, while the decoder can transform the extracted descriptor into another model.   {The SDN is able to learn to deform a template shape to be aligned to targets with the articulated restriction.   Concretely,  the encoder SDN learns the deformation parameters and degrees of freedom to deform the template.   This work shows that an encoder--decoder architecture to generate human shape correspondences can compete with state-of-the-art methods.}

 Convolutional Neuronal Networks have also been used for feature extraction.   {\mbox{\citet{Hanocka2018}} propose ALIGNet, an unsupervised network to align either 2D or 3D shapes with an incomplete target.   The network learns to extract the features to match both shapes and compute Free Form Deformation (FFD) grids.   It is trained with a shape alignment loss by comparing the overlap between the source and the target for learning the FFD parameters.} \citet{Ofir2018} developed a learning-based method to register multi-spectral images (visible and Near-Infra-Red images).   They employed a learning approach for extracting features of both images and matching them.   For that purpose, their proposal is based on an asymmetric {(different weights)} Siamese Convolutional Neural Network, one for each spectral channel.   The networks minimize the Euclidean distance between the two descriptors.   With a similar network architecture, \citet{Yew2018} proposed 3DFet-Net,  which  finds features and descriptors as well as correspondences for later registration.   They used  coarsely annotated point clouds with GPS/INS absolute pose.   It is based on a three-branch Siamese architecture that uses PointNet++ \cite{Qi2017a}.   {
Each branch takes an entire point cloud as input.   The network is trained with a set of triplets containing the anchor and  positive and negative point clouds.   Positive point clouds are those with a distance to the anchor below a threshold, and negative point clouds are far away from the anchor.   Each branch has a detector and descriptor network.   Both networks for each branch share the same inputs.   The detector network predicts an orientation and an attention weight for each branch.   Then, the descriptor network rotates the input to a canonical configuration and computes the features that will be aligned with the other branch through a triplet loss.   That loss aims to minimize the difference between the anchor and positive point cloud and maximize the difference between the anchor and the negative point cloud.
}

To address the problem of inaccurate correspondences, \citet{Schaffert2019} employed  a modified PointNet \cite{Charles2017} architecture for weighting individual correspondences in a 2D/3D rigid registration process on X-ray images.   {They employed  a modified PointNet to process points individually to obtain global information.   The authors included  a second MLP which processes correspondences containing global and local information.   This modified network is able to weight individual correspondences based on their geometrical properties and similarity as well as global properties.}

{
\citet{Chang2019} presented a 3D point set registration framework with two stages to cover the problem of coarse-to-fine registration.   Two descriptors are proposed, one for rough and one for fine orientation extraction,   SSPD and   8CBCP, respectively.     SSPD that is a normalized voxel grid and   8CBCP describes the orientation using an 8dx3 matrix obtained from the data in the eight  sub-quadrants of the bounding box around the 3D points.   They used two consecutive CNNs using those descriptors.   The first CNN receives as input the SSPD descriptor and is meant to estimate the coarse alignment.   Then, the 8CBCP descriptor is computed over the output and introduced in a second CCN that performs a more accurate alignment.   In this proposal, the CNNs only estimate the rotation, and  the translation is afterward obtained.   
}

Convolutions used in most of the deep learning networks operate over a neighborhood of the data; thus, structured inputs are required.   3D point clouds are unorganized datasets that are challenging to operate by convolution-based networks, a problem that led to much research on this topic.   Some state-of-the-art proposals tackle this problem by voxeling the point cloud \cite{Maturana2015} but these approaches are not efficient since points are sparse and a large percentage of voxels are empty and details can be lost.   Others try to extract geometric features from point clouds, e.g. \citet{Xu2018}   used   so-called SpiderConv filters that are parameterized functions of specific radius applied over the point cloud.   The ELF-Nets of \citet{Lee2019} proposed the Extended Laplacian Filter that is a combination of a two-state filter, one for the center point  and one for neighbors, with a scalar weighting function that represents the relative importance of the points.   This last approach uses fewer parameters than SpiderConvs.   For managing 3D data, \citet{Zeng2017} employed  a reduced set of voxels of TDF (truncate distance function) containing an interest point of a point cloud.   The TDF is the distance from the center of the voxel to the nearest point.   This is used as input of a convolutional network which extracts a 512-dimensional feature representation.   {The result is a geometric descriptor in which the network is able to generalize to other tasks and resolutions}.   However, according to \citet{Liu2018}, the CNN does not provide good results when working with point clouds due to their irregular structure.   For this purpose, they employed a modified version of the PointNet++ \cite{Qi2017a} architecture, a network that learns hierarchical features.   With that network, they propose a {network architecture that learns to predict scene flow as translational motion vectors for each point.}
{The proposed architecture has three modules: point feature learning, point mixture, and flow refinement.   It includes a \textit{flow embedding layer} that learns to aggregate geometric similarities and spatial relations of points for motion \mbox{encoding.   \citet{Pais2019}}} presented a network architecture with two main blocks, the classification block fed with pairs of corresponding 3D points and giving as a result features for each correspondence using 12 ResNets \cite{He2016},  which remove outlier correspondences.   The registration block gets the resulting features from the previous stage and produces a six-variable output for rotation and translation obtained with a context normalization layer along with convolutional one and two fully connected layers.   {This method works on point correspondences.   It is efficient and outperforms traditional approaches.}


\subsection{Transformation Level}

Some works have been successfully employed neural networks for learning and applying geometric transformations.   Some of these achievements have been done using GAN architectures or variants of them.   For instance, \citet{Lin2018} demonstrated good results using neural networks for finding realistic geometric transformations for 2D image compositing.   {Image compositing refers to overlap images coming from different scenes;  thus, achieving a good realism implies a good transformation to minimize the appearance and geometric differences.} For this purpose, they propose a GAN architecture using Spatial Transform Networks (STNs) \cite{jaderberg2015spatial}, named ST-GAN.   According to the authors of ST-GAN, this idea could be extended to other image alignment tasks.   {STNs have shown good results resolving geometric variations; thus,  with this architecture, the network learns to perform realistic geometric warps, demonstrating potential 3D capabilities.} \citet{Yan2018} also   used  GANs to carry out the registration of magnetic resonance (MR) and transrectal ultrasound (TRUS) images  as well as    evaluate the provided result.   The generator network provides the transformation parameters to align both inputs, while  the discriminator performs a quality evaluation of that alignment.   \citet{Mahapatra2018} used GANs for deformable multimodal medical image registration in 2D.   {The network outputs a transformed image and also a deformation field.} Similarly, in three-dimensional space, \citet{Hermoza2018}, {as  referenced above,} performed  the reconstruction of incomplete archaeological models also using GANs, {in which the generator network provides a reconstructed model.} {\citet{zhang2020deform} also proposed a registration method based on a GAN architecture with a gradient loss which can manage local structure information across modalities.   This makes it more robust against large deformations, noise, and blur.}

\citet{Ding2018} managed multiple point clouds registration using DNNs.   They approached  this problem by including two networks, a localization network named L-Net, and an occupancy map network, M-Net.   The L-Net estimates the sensor pose for a given point cloud, sharing some optimization parameters between the input point clouds.   The goal of this network is to estimate the sensor pose in a global frame.   To do that, the L-Net network is divided into a feature extraction module followed by an MLP that outputs the sensor pose.   The feature extraction module employed depends on the input format of the point cloud.   If it is an organized point cloud, a CNN is employed for that purpose.   If not, the features are extracted using PointNet \cite{Charles2017}.   Later, the M-Net receives those location coordinates in the global space and retrieves the discrete occupancy map.   Besides, the L-Net network locates each input point cloud in a global space;  there is no target for the alignment.
With a similar architecture, \citet{wang20193dn} presented 3DN, a combination of PointNet and MLPs that deforms 3D meshes to resemble a {target, given in the form of a 2D image or point cloud, as close as possible preserving the properties of the source.} The proposal extracts global features from both source and target inputs using CNN/PointNet.   Next, those features are used to estimate the per-vertex displacement with an ‘offset decoder’.   To overcome the problem of tessellation differences, an intermediate sampled point cloud is calculated from both source and target.   {They employed  a combination of four different loss functions, measuring the similarity between the deformed source and the target, symmetry, local geometric details, and self-intersections.   This work proposes an end-to-end network architecture for mesh deformation.}

Using autoencoders, \citet{Groueix2018}, with their SDNs introduced before, replicated  the shape of a body previously encoded in a given template.   {For 3D medical image non-rigid registration, \citet{Kuang2018} employed  an architecture inspired by STNs extending the works of \citet{shan2017unsupervised} and \citet{Balakrishnan2018}.   The network takes a pair of volumes and predicts the displacement fields needed to register source to target.   According to the authors, it improves the results compared to U-net \cite{Ronneberger2015} and VoxelMorph \cite{Balakrishnan2018}.   This method produces deformations with fewer regions of non-invertibility where the surface folds over itself.   To achieve this, they employed  an explicit anti-folding regularization to penalize \textit{foldings}, which are the spatial locations where the deformation is non-invertible and is indicated by a negative determinant of the Jacobian matrix.}

With convolutional networks, \citet{Jack2019} {performed  the 3D reconstruction from a single 2D image by learning to apply large deformations and compelling mesh reconstructions by inferring Free Form Deformation (FFD) parameters.   They employed  a lightweight CNN based on the MobileNet architecture \cite{howard2017mobilenets} to infer FDD parameters to deform a template and infer a 3D mesh of the given image.   As a result, the network learns how to deform a given template to match features present in a 2D image with finer geometry than other methods working with voxel grids and point clouds, because there is no discretization.}

\citet{Guan2019} proposed a multi-channel CNN (MCNN) for deformable registration of CT scans {with digital subtraction angiography (DSA) images of the cardiovascular system.   The network is composed of several sub-networks that converge before the fully connected layers.   They named  this architecture as a multi-channel convolutional neural network.   They employed  a CNN model based on the VGG network combined with a vascular diameter variation model to directly regress and predict transformation parameters.   With this architecture, each channel of the MCNN process a different phase of the vascular deformation cycle, comparing the results of each to choose the best result.} \citet{Li2017} employed  Fully Convolutional Networks (FCNs) to optimize and learn spatial transformations between pairs of images to be non-rigidly registered.   Their method works with medical images at the voxel level and, according to the authors, it improves the results of STNs, which cannot manage small transformations.   {The spatial transformation between pairs of images is obtained directly by maximizing an image-wise similarity metric, similar to traditional approaches.   The use of FCNs facilitates the voxel-to-voxel prediction of deformation fields, which also allows learning small transformations.}

\citet{Gundogdu2018} proposed a method for 3D garment fitting on bodies.   {To extract global features of the body model,  they employed a PointNet \cite{Charles2017} but with leaky  ReLUs with a slope of $0.1$.} {After that, a second stream, composed by six residual blocks, is used to extract features from the garment mesh and also take as input the previous global body features.} Thirdly, the features provided by both networks are merged employing {four Multi Layer Perceptron (MLP) blocks shared by all points.   The final MLP block outputs a vector with the 3D translation information.} With this method, the authors achieved  results nearly as accurate as a Physics-Based Simulation (PBS), but less time-consuming.

Similarly, PointNetLK \cite{Aoki2019} is a method for 3D rigid registration {which modifies the Lucas  and  Kanade (LK) \cite{lucas1981iterative} algorithm integrated with PointNet.   The process is mainly} divided into two steps: initially, two 3D point sets are passed through a shared Multi Layer Perceptron of the two inputs and a symmetric pooling function.   Second, the transformation is obtained and applied to the moving point cloud.   The whole procedure is iteratively repeated until a minimum quality threshold is reached.   {According to the authors, this method exhibits remarkable generalization of unseen objects and shape variation due to the encoding of the alignment process in the network architecture that only needs to learn the PointNet representation.} {\citet{li2020deterministic} proposed  a deterministic PointNetLK method to improve the generalization by using analytical gradients. }
\citet{wang2019coherent} presented CPD-Net, a network architecture that performs non-rigid registration under the concept of {learning a displacement vector function that estimates the geometric transformation}.   The pipeline is decomposed into three main components: ‘Learning Shape Descriptor’ with a MultiLayer Perceptron (MLP) that learns descriptors from the input source and target point sets; ‘Coherent PointMorph’ that is a three MLPs block fed with the two descriptors concatenated with the source data points;  and  the ‘Point Set Alignment’, {where the loss function is defined to determine the quality of the alignment.} Deep Closest Point \cite{Wang_2019_ICCV} registers two point clouds by first embedding them into high-dimensional space using DGCNN \cite{Wang2019} to extract features.   After that, contextual information is estimated using an attention-based module that provides a dependency term between the feature sets, i.e., one set is modified in a way that is knowledgeable about the structure of the other.   Finally, alignment is obtained using a differentiable Singular Value Decomposition (SVD) layer, which seems to provide better results than an MLP.   This proposal also includes a ``pointer generation'' that {provides a probabilistic approach to generate a soft map between the two features sets to minimize the problem of falling into a local minima}.

Going back to the work of \citet{Pais2019} cited  above,  the component performing the alignment uses a CNN that receives as input features extracted from the selected correspondences at different stages of the previous components composed by ResNet blocks.   {The registration block receives as input features previously extracted and outputs the transformation parameters.   }


\subsection{Summary}

Through this section, an overview of neural networks that perform alignment or registration tasks  is provided.   A perspective based on the traditionally employed pipeline in registration methods   is used to analyze the proposals.   From the analyzed works (summarized in Table \ref{tab:methods}), it is possible to observe that the extraction and matching of features are tasks widely explored with neural networks because they are common points with other problems  such as  object classification or recognition.   However, it is not common to find neural networks with the ability to manage geometric information for applying transformations on data to meet some requirements.

There are approaches dealing with some parts of the pipeline  or the whole registration.   Interestingly, the proposals that compute the transformation for the alignment also perform the matching of features.   That means there are no proposals performing only the calculation of the transformation.   In terms of deformable alignment, there exist specific proposals for learning deformations or non-rigid registration with networks,  such as  the SDNs, but is a current topic under research.

In addition, it is noticeable that most of the analyzed neural networks employ a GAN, a CNN, or a variation of them, but there are a few works with a GAN architecture managing 3D data.   However, this point is under active study because this kind of data requires many resources.   As occurs with 2D images, the main solution is to use discrete input data.   Similar to pixels in 2D, voxel grids are the common solution for 3D.   However, in some situations, it is important to work at point level, e.g.,    for estimating deformation flow.   Although there are some proposals working with point clouds as input data, most of them require an organized point cloud or a limitation in the number of points.

{
	From the reviewed methods,  it is possible to extract the key innovations that are relevant to registration problems.   Table \ref{tab:summary} summarizes the key contributions of the reviewed methods classified according to the stage in which they are relevant.   

}

\newcommand{\tabitem}{ \llap{\textbullet} }

\newcommand{\tabitems}{\hspace*{3em} \llap{\textbullet} }

\begin{table}[!t]
		
		

	\centering
	\caption{Summary of the key advantages of the reviewed methods.}
	\label{tab:summary}
	{\small
		\scalebox{.9}[0.9]{
		{
\begin{tabularx}{\textwidth}{p{3em}X}
	\midrule
	\multicolumn{2}{X}{
	\textbf{\textit{target selection}} }\\
	\midrule

	& \vspace{-6pt}
	\begin{itemize}[labelsep=2mm]
	\item Reconstruction of 3D models from a single 2D image using encoder--decoder architectures \cite{Wang2017,Smirnov2019}
	\item Leverage partial 3D observations to generate complete 3D models (mesh completion) using Generative Adversarial Networks (GANs) and Variational Autoencoders (VAEs) for predicting missing geometry \cite{Hermoza2018, Litany2018}
	\item The use of Adversarial Training (AT) to predict structural deformations on 3D meshes given a multi-modal input (e.g., forces, material properties, visual imaging) \cite{Wang2018b}
	 \end{itemize}\\
	\midrule

	\multicolumn{2}{X}{
		\textbf{\textit{features and matching}} }\\
	\midrule

	& \vspace{-6pt}
	\begin{itemize}[labelsep=2mm]
	\item Registration of global and local point clouds with a generic Deep Auto Encoder (DAE) architecture regardless of the input data and its source \cite{Elbaz2017}
	\item The Shape Deformation Network (SDN) is an autoencoder architecture able to deal with deformable shapes extracting global shape descriptors \cite{Groueix2018}
	\item Capability for multi-spectral registration using Asymmetric Siamese Convolutional Networks \cite{Ofir2018}
	\item Filtering inaccurate features correspondences based on geometrical and global properties to minimize their influence in the registration process \cite{Schaffert2019,Pais2019}
	\item Registration process speedup using a two-staged approach based on Convolutional Neural Networks (CNNs) to solve coarse-to-fine registration problems \cite{Chang2019}
	\end{itemize}\\
	\midrule

	\multicolumn{2}{X}{
		\textbf{\textit{transform}} }\\
	\midrule

	& \vspace{-6pt}
	\begin{itemize}[labelsep=2mm]
	\item Using Spatial Transformer Networks (STNs) to deal with geometrical variability by means of learned invariance to transformations \cite{jaderberg2015spatial,Lin2018,Kuang2018}
	\item To align multimodal inputs with transformations generated and evaluated in an adversarial fashion \cite{Yan2018}
	\item Ability to preserve detailed geometric information by using a CNN to infer Free Form Deformation (FFD) parameters for 3D template-image matching \cite{Jack2019}
	\item Unsupervised registration of multiple point clouds in a global frame of reference \cite{Ding2018}
	\item Outperforming single channel CNNs with a multi-channel approach for real-time deformable registration \cite{Guan2019}
	\item Cloth fitting by modeling 3D body-garment interaction in real time surpassing Physics-Based Simulation (PBS) \cite{Gundogdu2018}
	\end{itemize}\\
	
	\bottomrule
\end{tabularx}}
}
}

\end{table}



\section{Conclusions}
\label{sec:discussion}

In this work, approaches in the intersection between registration and deep neural networks   are  reviewed.   It is important to remark that a large part of the works  was  reached through the ArXiv repository, which at the moment, even though is not peer-reviewed, leads the advancements in many fields   such as   artificial intelligence and deep learning.   For this reason,   extra effort was required to select the papers.   However, this approach enables including the most recent works in the scope of this paper.

Registration aims to calculate a common reference for two or more     datasets.   This field has been widely studied, but recent techniques in machine learning are being combined with registration algorithms to increase the capabilities of the proposals.   These techniques include neural networks with many hidden layers, also known as deep learning, and its novelty remains in the ability to learn representations from huge amounts of data at different levels of abstraction.{
	Applied to registration, this paradigm allows managing higher-level understanding problems that are more related to conceptual knowledge of the scene rather than the geometric properties.   We name this paradigm Deep Registration Networks (DRNs) to identify the branch of artificial intelligence exploring solutions for alignment problems using DNNs.}

The contribution of this work is a review of registration methods based on deep networks.   To       do that, the learning approaches for registration   are reviewed and classified using a novel framework extracted from the traditional registration pipeline.   The review clearly identifies the current efforts and existing gaps in the intersection between registration and learning algorithms.   Moreover, the positive influence of the internal representations modeled by learning approaches on the registration process  is clearly identified.


 As a result, an overall view of this new subfield is provided, setting out different architectures and solutions that are being provided by the authors.   A summary of the different methods is shown in Table \ref{tab:methods} with the inputs, outputs, architectures, and datasets employed to address the problems.   Besides, an analysis of each method has been made using a traditional perspective of the pipeline employed in registration algorithms.   As the main contribution of this work, we provide a framework to understand the learning methods for registration.   In these new methods, the stages of the pipeline are not so clearly defined as they are in a traditional approach, because some processes are computed directly and implicitly by the network, e.g.,    the extraction and matching of features.   However, an advantage of the learning approaches is that they are suitable for real-time problems.   The higher computational needs of a neural network are at the training phase, which is performed once.   After that, the data processing is relatively fast for real-time applications.
 
 
 From our perspective, it is clear that researchers are still exploring different paradigms, and no single approach is so far the preferred one.   Whether the learning-based approaches will enable significant improvements over traditional registration approaches is still an open question.   To help assess whether convergence in the literature is happening, we analyzed the approaches using k-means and SOM networks to find clusters of methods sharing characteristics.   However, no significant clusters were found, suggesting that convergence has not yet happened.   {The metrics employed to evaluate each method are different, some of them are even ad hoc solutions.   In addition, there is a lack of standard benchmarks as well as common datasets to compare/evaluate the methods.   For this reason, a comparison between methods is  not   a contribution as it  would  not allow extracting relevant conclusions.}
 
To conclude, we find that most current approaches can be analyzed using concepts from the four stages of registration  identified in Figure \ref{fig1}, which enable the recognition, registration, and reconstruction of objects.   Although the four stages  are  evident in the traditional algorithms, with the rise of deep learning, we believe that it will be possible to deal with more complex registration problems, e.g., at the conceptual level.
 

\vspace{6pt}


\funding{{This work  was  supported by the Spanish State Research Agency (AEI) and the European Regional Development Fund (FEDER) under project TIN2017-89069-R.   This work  was also   supported by a Valencian Regional project (GV/2020/056), two Valencian Grants for Ph.D.   studies (ACIF/2017/223 and ACIF/2018/197), and two Valencian Grants for predoctoral internships (BEFPI/2020/001 and BEFPI/2020/068).}} 






\begin{thebibliography}{999}

\bibitem[Villena-Martinez et al.(2017)Villena-Martinez, Fuster-Guillo,
 Saval-Calvo, and Azorin-Lopez]{VillenaMartinez2017}
Villena-Martinez, V.; Fuster-Guillo, A.; Saval-Calvo, M.; Azorin-Lopez, J.
\newblock 3D Body Registration from {RGB}-D Data with Unconstrained Movements
 and Single Sensor.   In {\em International Work-Conference on Artificial Neural Networks}; Springer International Publishing: Cham, Switzerland, 2017;
 pp.   317--329.
\newblock
 doi:{\changeurlcolor{black}\href{https://doi.org/10.1007/978-3-319-59147-6_28}{\detokenize{10.1007/978-3-319-59147-6_28}}}.

\bibitem[Zeman et al.(1994)Zeman, Davros, Berman, Weltman, Silverman,
 Cooper, Evans, Buras, Stahl, and Nauta]{Zeman1994}
Zeman, R.K.; Davros, W.J.; Berman, P.; Weltman, D.I.; Silverman, P.M.; Cooper,
 C.; Evans, S.R.; Buras, R.R.; Stahl, T.J.; Nauta, R.J.
\newblock Three-dimensional models of the abdominal vasculature based on
 helical {CT}: usefulness in patients with pancreatic neoplasms.
\newblock {\em Am.   J.   Roentgenol.} {\bf 1994}, {\em 162}, 1425--1429.
\newblock
 doi:{\changeurlcolor{black}\href{https://doi.org/10.2214/ajr.162.6.8192012}{\detokenize{10.2214/ajr.162.6.8192012}}}.

\bibitem[Boldea et al.(2008)Boldea, Sharp, Jiang, and Sarrut]{Boldea2008}
Boldea, V.; Sharp, G.C.; Jiang, S.B.; Sarrut, D.
\newblock 4D-{CT} lung motion estimation with deformable registration:
 Quantification of motion nonlinearity and hysteresis.
\newblock {\em Med.   Phys.} {\bf 2008}, {\em 35}, 1008--1018.
\newblock
 doi:{\changeurlcolor{black}\href{https://doi.org/10.1118/1.2839103}{\detokenize{10.1118/1.2839103}}}.

\bibitem[Cuevas-Velasquez et al.(2018)Cuevas-Velasquez, Li, Tylecek,
 Saval-Calvo, and Fisher]{CuevasVelasquez2018}
Cuevas-Velasquez, H.; Li, N.; Tylecek, R.; Saval-Calvo, M.; Fisher, R.B.
\newblock Hybrid Multi-camera Visual Servoing to Moving Target.
\newblock In Proceedings of the 2018 IEEE/RSJ International Conference on Intelligent Robots and Systems (IROS), Madrid, Spain, 1--5 October 2018.
\newblock
 doi:{\changeurlcolor{black}\href{https://doi.org/10.1109/iros.2018.8593652}{\detokenize{10.1109/iros.2018.8593652}}}.

\bibitem[Zhao et al.(2019)Zhao, Zheng, Xu, and Wu]{Zhao2019}
Zhao, Z.Q.; Zheng, P.; Xu, S.T.; Wu, X.
\newblock Object Detection With Deep Learning: A Review.
\newblock {\em {IEEE} Trans.   Neural Netw.   Learn.   Syst.} {\bf 2019}, {\em
 30}, 3212--3232.
\newblock
 doi:{\changeurlcolor{black}\href{https://doi.org/10.1109/tnnls.2018.2876865}{\detokenize{10.1109/tnnls.2018.2876865}}}.

\bibitem[Garcia-Garcia et al.(2018)Garcia-Garcia, Orts-Escolano, Oprea,
 Villena-Martinez, Martinez-Gonzalez, and Garcia-Rodriguez]{Garcia-Garcia2018}
Garcia-Garcia, A.; Orts-Escolano, S.; Oprea, S.; Villena-Martinez, V.;
 Martinez-Gonzalez, P.; Garcia-Rodriguez, J.
\newblock A survey on deep learning techniques for image and video semantic
 segmentation.
\newblock {\em Appl.   Soft Comput.} {\bf 2018}, {\em 70}, 41--65.
\newblock
 doi:{\changeurlcolor{black}\href{https://doi.org/10.1016/j.asoc.2018.05.018}{\detokenize{10.1016/j.asoc.2018.05.018}}}.

\bibitem[Oprea et al.(2020)Oprea, Martinez-Gonzalez, Garcia-Garcia,
 Castro-Vargas, Orts-Escolano, Garcia-Rodriguez, and Argyros]{oprea2020review}
Oprea, S.; Martinez-Gonzalez, P.; Garcia-Garcia, A.; Castro-Vargas, J.A.;
 Orts-Escolano, S.; Garcia-Rodriguez, J.; Argyros, A.
\newblock A Review on Deep Learning Techniques for Video Prediction.
\newblock {\em arXiv} {\bf 2020}, arXiv:2004.05214.

\bibitem[Tam et al.(2013)Tam, Cheng, Lai, Langbein, Liu, Marshall, Martin,
 Sun, and Rosin]{Tam2013}
Tam, G.K.L.; Cheng, Z.Q.; Lai, Y.K.; Langbein, F.C.; Liu, Y.; Marshall, D.;
 Martin, R.R.; Sun, X.F.; Rosin, P.L.
\newblock Registration of 3D Point Clouds and Meshes: A Survey from Rigid to
 Nonrigid.
\newblock {\em {IEEE} Trans.   Vis.   Comput.   Graph.} {\bf 2013}, {\em
 19}, 1199--1217.
\newblock
 doi:{\changeurlcolor{black}\href{https://doi.org/10.1109/tvcg.2012.310}{\detokenize{10.1109/tvcg.2012.310}}}.

\bibitem[Zhu et al.(2019)Zhu, Guo, Zou, Li, Yuen, Mihaylova, and
 Leung]{Zhu2019}
Zhu, H.; Guo, B.; Zou, K.; Li, Y.; Yuen, K.V.; Mihaylova, L.; Leung, H.
\newblock A Review of Point Set Registration:From Pairwise Registration to
 Groupwise Registration.
\newblock {\em Sensors} {\bf 2019}, {\em 19}, 1191.
\newblock
 doi:{\changeurlcolor{black}\href{https://doi.org/10.3390/s19051191}{\detokenize{10.3390/s19051191}}}.

\bibitem[Salvi et al.(2007)Salvi, Matabosch, Fofi, and Forest]{Salvi2007}
Salvi, J.; Matabosch, C.; Fofi, D.; Forest, J.
\newblock A review of recent range image registration methods with accuracy
 evaluation.
\newblock {\em Image Vis.   Comput.} {\bf 2007}, {\em 25}, 578--596.
\newblock
 doi:{\changeurlcolor{black}\href{https://doi.org/10.1016/j.imavis.2006.05.012}{\detokenize{10.1016/j.imavis.2006.05.012}}}.
 
 
 \bibitem[Yumer and Mitra(2016)]{Yumer2016}
Yumer, M.E.; Mitra, N.J.
\newblock Learning Semantic Deformation Flows with 3D Convolutional Networks.
 In {\em {ECCV}}; Springer: Berlin/Heidelberg, Germany, 2016; pp.   294--311.
\newblock
 doi:{\changeurlcolor{black}\href{https://doi.org/10.1007/978-3-319-46466-4_18}{\detokenize{10.1007/978-3-319-46466-4_18}}}.

\bibitem[Chang et al.(2015)Chang, Funkhouser, Guibas, Hanrahan, Huang, Li,
 Savarese, Savva, Song, Su, et al.]{chang2015shapenet}
Chang, A.X.; Funkhouser, T.; Guibas, L.; Hanrahan, P.; Huang, Q.; Li, Z.;
 Savarese, S.; Savva, M.; Song, S.; Su, H.; et al.
\newblock Shapenet: An information-rich 3d model repository.
\newblock {\em arXiv} {\bf 2015}, arXiv:1512.03012.

\bibitem[Yumer et al.(2015)Yumer, Chaudhuri, Hodgins, and Kara]{Yumer2015}
Yumer, M.E.; Chaudhuri, S.; Hodgins, J.K.; Kara, L.B.
\newblock Semantic shape editing using deformation handles.
\newblock {\em {ACM} Trans.   Graph.} {\bf 2015}, {\em 34}, 1--12.
\newblock
 doi:{\changeurlcolor{black}\href{https://doi.org/10.1145/2766908}{\detokenize{10.1145/2766908}}}.

\bibitem[Elbaz et al.(2017)Elbaz, Avraham, and Fischer]{Elbaz2017}
Elbaz, G.; Avraham, T.; Fischer, A.
\newblock 3D Point Cloud Registration for Localization Using a Deep Neural
 Network Auto-Encoder.
\newblock In Proceedings of the IEEE Conference on Computer Vision and Pattern Recognition, Honolulu, HI, USA, 21--26 July 2017.
\newblock
 doi:{\changeurlcolor{black}\href{https://doi.org/10.1109/cvpr.2017.265}{\detokenize{10.1109/cvpr.2017.265}}}.

\bibitem[Pomerleau et al.(2012)Pomerleau, Liu, Colas, and
 Siegwart]{Pomerleau2012}
Pomerleau, F.; Liu, M.; Colas, F.; Siegwart, R.
\newblock Challenging data sets for point cloud registration algorithms.
\newblock {\em Int.   J.   Rob.   Res.} {\bf 2012}, {\em 31}, 1705--1711.
\newblock
 doi:{\changeurlcolor{black}\href{https://doi.org/10.1177/0278364912458814}{\detokenize{10.1177/0278364912458814}}}.

\bibitem[Li and Fan(2017)]{Li2017}
Li, H.; Fan, Y.
\newblock Non-rigid image registration using fully convolutional networks with
 deep self-supervision.
\newblock {\em arXiv} {\bf 2017}, arXiv:1709.00799.

\bibitem[Wang and Fang(2017)]{Wang2017}
Wang, L.; Fang, Y.
\newblock Unsupervised 3D reconstruction from a single image via adversarial
 learning.
\newblock {\em arXiv} {\bf 2017}, arXiv:1711.09312.

\bibitem[Xiang et al.(2014)Xiang, Mottaghi, and Savarese]{Xiang2014}
Xiang, Y.; Mottaghi, R.; Savarese, S.
\newblock Beyond {PASCAL}: A benchmark for 3D object detection in the wild.
\newblock In Proceedings of the IEEE Winter Conference on Applications of Computer Vision, Steamboat Springs, CO, USA, 24--26 March 2014.
\newblock
 doi:{\changeurlcolor{black}\href{https://doi.org/10.1109/wacv.2014.6836101}{\detokenize{10.1109/wacv.2014.6836101}}}.

\bibitem[Li et al.(2013)Li, Lu, Godil, Schreck, Aono, Johan, Saavedra, and
 Tashiro]{li2013shrec}
Li, B.; Lu, Y.; Godil, A.; Schreck, T.; Aono, M.; Johan, H.; Saavedra, J.M.;
 Tashiro, S.
\newblock SHREC'13 Track: Large Scale Sketch-Based 3D Shape Retrieval.
\newblock \emph{EG 3DOR}; Eurographics: Aire-la-Ville, Switzerland, 2013, pp.   89--96.
\newblock
 doi:{\changeurlcolor{black}\href{https://doi.org/10.2312/3DOR/3DOR13/089-096}{\detokenize{10.2312/3DOR/3DOR13/089-096}}}.

\bibitem[Zeng et al.(2017)Zeng, Song, NieBner, Fisher, Xiao, and
 Funkhouser]{Zeng2017}
Zeng, A.; Song, S.; NieBner, M.; Fisher, M.; Xiao, J.; Funkhouser, T.
\newblock 3DMatch: Learning Local Geometric Descriptors from {RGB}-D
 Reconstructions.
\newblock In Proceedings of the IEEE Conference on Computer Vision and Pattern Recognition (CVPR), Honolulu, HI, USA, 21--26 July 2017.
\newblock
 doi:{\changeurlcolor{black}\href{https://doi.org/10.1109/cvpr.2017.29}{\detokenize{10.1109/cvpr.2017.29}}}.

\bibitem[Valentin et al.(2016)Valentin, Dai, Niessner, Kohli, Torr, Izadi,
 and Keskin]{Valentin2016}
Valentin, J.; Dai, A.; Niessner, M.; Kohli, P.; Torr, P.; Izadi, S.; Keskin, C.
\newblock Learning to Navigate the Energy Landscape.
\newblock In Proceedings of the 2016 Fourth International Conference on 3D Vision (3DV), Stanford, CA, USA, 25--28 October 2016.
\newblock
 doi:{\changeurlcolor{black}\href{https://doi.org/10.1109/3dv.2016.41}{\detokenize{10.1109/3dv.2016.41}}}.

\bibitem[Shotton et al.(2013)Shotton, Glocker, Zach, Izadi, Criminisi, and
 Fitzgibbon]{Shotton2013}
Shotton, J.; Glocker, B.; Zach, C.; Izadi, S.; Criminisi, A.; Fitzgibbon, A.
\newblock Scene Coordinate Regression Forests for Camera Relocalization in
 {RGB}-D Images.
\newblock In Proceedings of the IEEE Conference on Computer Vision and Pattern Recognition (CVPR), Portland, OR, USA, 23--28 June 2013.
\newblock
 doi:{\changeurlcolor{black}\href{https://doi.org/10.1109/cvpr.2013.377}{\detokenize{10.1109/cvpr.2013.377}}}.

\bibitem[Xiao et al.(2013)Xiao, Owens, and Torralba]{Xiao2013}
Xiao, J.; Owens, A.; Torralba, A.
\newblock {SUN}3D: A Database of Big Spaces Reconstructed Using {SfM} and
 Object Labels.
\newblock In Proceedings of the IEEE International Conference on Computer Vision (ICCV), Sydney, Australia, 1--8 December 2013.
\newblock
 doi:{\changeurlcolor{black}\href{https://doi.org/10.1109/iccv.2013.458}{\detokenize{10.1109/iccv.2013.458}}}.
 
 \bibitem[Lai et al.(2014)Lai, Bo, and Fox]{Lai2014}
Lai, K.; Bo, L.; Fox, D.
\newblock Unsupervised feature learning for 3D scene labeling.
\newblock In Proceedings of the 2014 IEEE International Conference on Robotics and Automation (ICRA), Hong Kong, China, 31 May--7 June 2014.
\newblock
 doi:{\changeurlcolor{black}\href{https://doi.org/10.1109/icra.2014.6907298}{\detokenize{10.1109/icra.2014.6907298}}}.

\bibitem[Halber and Funkhouser(2017)]{Halber2017}
Halber, M.; Funkhouser, T.
\newblock Fine-to-Coarse Global Registration of {RGB}-D Scans.
\newblock In Proceedings of the IEEE Conference on Computer Vision and Pattern Recognition (CVPR), Honolulu, HI, USA, 21--26 July 2017.
\newblock
 doi:{\changeurlcolor{black}\href{https://doi.org/10.1109/cvpr.2017.705}{\detokenize{10.1109/cvpr.2017.705}}}.

\bibitem[Ding and Feng(2019)]{Ding2018}
Ding, L.; Feng, C.
\newblock {DeepMapping}: Unsupervised Map Estimation From Multiple Point
 Clouds.
\newblock In Proceedings of the IEEE/CVF Conference on Computer Vision and Pattern Recognition (CVPR), Long Beach, CA, USA, 15--20 June 2019.
\newblock
 doi:{\changeurlcolor{black}\href{https://doi.org/10.1109/cvpr.2019.00885}{\detokenize{10.1109/cvpr.2019.00885}}}.

\bibitem[Ammirato et al.(2017)Ammirato, Poirson, Park, Kosecka, and
 Berg]{Ammirato2017}
Ammirato, P.; Poirson, P.; Park, E.; Kosecka, J.; Berg, A.C.
\newblock A dataset for developing and benchmarking active vision.
\newblock In Proceedings of the 2017 IEEE International Conference on Robotics and Automation (ICRA), Singapore, 29 May--3 June 2017.
\newblock
 doi:{\changeurlcolor{black}\href{https://doi.org/10.1109/icra.2017.7989164}{\detokenize{10.1109/icra.2017.7989164}}}.

\bibitem[Groueix et al.(2018)Groueix, Fisher, Kim, Russell, and
 Aubry]{Groueix2018}
Groueix, T.; Fisher, M.; Kim, V.G.; Russell, B.C.; Aubry, M.
\newblock 3D-{CODED}: 3D Correspondences by Deep Deformation.   In {\em {ECCV}};
 Springer International Publishing: Cham, Switzerland, 2018; pp.   235--251.
\newblock
 doi:{\changeurlcolor{black}\href{https://doi.org/10.1007/978-3-030-01216-8_15}{\detokenize{10.1007/978-3-030-01216-8_15}}}.

\bibitem[Loper et al.(2015)Loper, Mahmood, Romero, Pons-Moll, and
 Black]{Loper2015}
Loper, M.; Mahmood, N.; Romero, J.; Pons-Moll, G.; Black, M.J.
\newblock {SMPL}.
\newblock {\em {ACM} Trans.   Graph.} {\bf 2015}, {\em 34}, 1--16.
\newblock
 doi:{\changeurlcolor{black}\href{https://doi.org/10.1145/2816795.2818013}{\detokenize{10.1145/2816795.2818013}}}.

\bibitem[Varol et al.(2017)Varol, Romero, Martin, Mahmood, Black, Laptev,
 and Schmid]{Varol2017}
Varol, G.; Romero, J.; Martin, X.; Mahmood, N.; Black, M.J.; Laptev, I.;
 Schmid, C.
\newblock Learning from Synthetic Humans.
\newblock In Proceedings of the IEEE Conference on Computer Vision and Pattern Recognition (CVPR), Honolulu, HI, USA, 21--26 July 2017.
\newblock
 doi:{\changeurlcolor{black}\href{https://doi.org/10.1109/cvpr.2017.492}{\detokenize{10.1109/cvpr.2017.492}}}.

\bibitem[Zuffi et al.(2017)Zuffi, Kanazawa, Jacobs, and Black]{Zuffi2017}
Zuffi, S.; Kanazawa, A.; Jacobs, D.W.; Black, M.J.
\newblock 3D Menagerie: Modeling the 3D Shape and Pose of Animals.
\newblock In Proceedings of the IEEE Conference on Computer Vision and Pattern Recognition (CVPR), Honolulu, HI, USA, 21--26 July 2017.
\newblock
 doi:{\changeurlcolor{black}\href{https://doi.org/10.1109/cvpr.2017.586}{\detokenize{10.1109/cvpr.2017.586}}}.

\bibitem[Bogo et al.(2014)Bogo, Romero, Loper, and Black]{Bogo2014}
Bogo, F.; Romero, J.; Loper, M.; Black, M.J.
\newblock {FAUST}: Dataset and Evaluation for 3D Mesh Registration.
\newblock In Proceedings of the IEEE Conference on Computer Vision and Pattern Recognition (CVPR), 24-27 June 2014, Columbus, OH, USA.
\newblock
 doi:{\changeurlcolor{black}\href{https://doi.org/10.1109/cvpr.2014.491}{\detokenize{10.1109/cvpr.2014.491}}}.

\bibitem[Bronstein et al.(2008)Bronstein, Bronstein, and
 Kimmel]{bronstein2008numerical}
Bronstein, A.M.; Bronstein, M.M.; Kimmel, R.
\newblock {\em Numerical Geometry of Non-Rigid Shapes}; Springer Science \&
 Business Media: New York, NY, USA, 2008.

\bibitem[Anguelov et al.(2005)Anguelov, Srinivasan, Koller, Thrun,
 Rodgers, and Davis]{Anguelov2005}
Anguelov, D.; Srinivasan, P.; Koller, D.; Thrun, S.; Rodgers, J.; Davis, J.
\newblock {SCAPE}.
\newblock \emph{{ACM} {SIGGRAPH} 2005}; {ACM} Press: New York, NY, USA, 2005.
\newblock
 doi:{\changeurlcolor{black}\href{https://doi.org/10.1145/1186822.1073207}{\detokenize{10.1145/1186822.1073207}}}.

\bibitem[Gundogdu et al.(2019)Gundogdu, Constantin, Seifoddini, Dang,
 Salzmann, and Fua]{Gundogdu2018}
Gundogdu, E.; Constantin, V.; Seifoddini, A.; Dang, M.; Salzmann, M.; Fua, P.
\newblock Garnet: A two-stream network for fast and accurate 3d cloth draping.
\newblock In Proceedings of the IEEE/CVF International Conference on Computer Vision (ICCV), Seoul, Korea, 27--28 October 2019; pp.   8739--8748.

\bibitem[Hanocka et al.(2018)Hanocka, Fish, Wang, Giryes, Fleishman, and
 Cohen-Or]{Hanocka2018}
Hanocka, R.; Fish, N.; Wang, Z.; Giryes, R.; Fleishman, S.; Cohen-Or, D.
\newblock {ALIGNet}.
\newblock {\em {ACM} Trans.   Graph.} {\bf 2018}, {\em 38}, 1--14.
\newblock
 doi:{\changeurlcolor{black}\href{https://doi.org/10.1145/3267347}{\detokenize{10.1145/3267347}}}.

\bibitem[Wang et al.(2012)Wang, Asafi, van Kaick, Zhang, Cohen-Or, and
 Chen]{Wang2012}
Wang, Y.; Asafi, S.; van Kaick, O.; Zhang, H.; Cohen-Or, D.; Chen, B.
\newblock Active co-analysis of a set of shapes.
\newblock {\em {ACM} Trans.   Graph.} {\bf 2012}, {\em 31}, 1.
\newblock
 doi:{\changeurlcolor{black}\href{https://doi.org/10.1145/2366145.2366184}{\detokenize{10.1145/2366145.2366184}}}.

\bibitem[Hermoza and Sipiran(2018)]{Hermoza2018}
Hermoza, R.; Sipiran, I.
\newblock 3D Reconstruction of Incomplete Archaeological Objects Using a
 Generative Adversarial Network.
\newblock In Proceedings of the Computer Graphics International 2018, Bintan Island, Indonesia, 11--14 June 2018.
\newblock
 doi:{\changeurlcolor{black}\href{https://doi.org/10.1145/3208159.3208173}{\detokenize{10.1145/3208159.3208173}}}.

\bibitem[Wu et al.(2015)Wu, Song, Khosla, Yu, Zhang, Tang, and
 Xiao]{ZhirongWu2015}
Wu, Z.; Song, S.; Khosla, A.; Yu, F.; Zhang, L.; Tang, X.; Xiao, J.
\newblock 3D {ShapeNets}: A deep representation for volumetric shapes.
\newblock In Proceedings of the IEEE Conference on Computer Vision and Pattern Recognition (CVPR), Boston, MA, USA, 7--12 June 2015.
\newblock
 doi:{\changeurlcolor{black}\href{https://doi.org/10.1109/cvpr.2015.7298801}{\detokenize{10.1109/cvpr.2015.7298801}}}.

\bibitem[Koutsoudis et al.(2010)Koutsoudis, Pavlidis, Liami, Tsiafakis,
 and Chamzas]{Koutsoudis2010}
Koutsoudis, A.; Pavlidis, G.; Liami, V.; Tsiafakis, D.; Chamzas, C.
\newblock 3D Pottery content-based retrieval based on pose normalisation and
 segmentation.
\newblock {\em J.   Cult.   Herit.} {\bf 2010}, {\em 11}, 329--338.
\newblock
 doi:{\changeurlcolor{black}\href{https://doi.org/10.1016/j.culher.2010.02.002}{\detokenize{10.1016/j.culher.2010.02.002}}}.

\bibitem[Yew and Lee(2018)]{Yew2018}
Yew, Z.J.; Lee, G.H.
\newblock 3DFeat-Net: Weakly Supervised Local 3D Features for Point Cloud
 Registration.   In {\em {ECCV}}; Springer: Berlin/Heidelberg, Germany, 2018; pp.   630--646.
\newblock
 doi:{\changeurlcolor{black}\href{https://doi.org/10.1007/978-3-030-01267-0_37}{\detokenize{10.1007/978-3-030-01267-0_37}}}.

\bibitem[Maddern et al.(2016)Maddern, Pascoe, Linegar, and
 Newman]{Maddern2016}
Maddern, W.; Pascoe, G.; Linegar, C.; Newman, P.
\newblock 1 year, 1000 km: The Oxford {RobotCar} dataset.
\newblock {\em \mbox{Int.   J.   Rob.   Res.}} {\bf 2016}, {\em 36}, 3--15.
\newblock
 doi:{\changeurlcolor{black}\href{https://doi.org/10.1177/0278364916679498}{\detokenize{10.1177/0278364916679498}}}.

\bibitem[Geiger et al.(2012)Geiger, Lenz, and Urtasun]{Geiger2012}
Geiger, A.; Lenz, P.; Urtasun, R.
\newblock Are we ready for autonomous driving? The {KITTI} vision benchmark
 suite.
\newblock In Proceedings of the 2012 IEEE Conference on Computer Vision and Pattern Recognition, Providence, RI, USA, 16--21 June 2012.   
\newblock
 doi:{\changeurlcolor{black}\href{https://doi.org/10.1109/cvpr.2012.6248074}{\detokenize{10.1109/cvpr.2012.6248074}}}.

\bibitem[Kuang and Schmah(2019)]{Kuang2018}
Kuang, D.; Schmah, T.
\newblock {FAIM} {\textendash} A {ConvNet} Method for Unsupervised 3D Medical
 Image Registration.   In {\em MLMI}; Springer: Berlin/Heidelberg, Germany, 2019; pp.   646--654.
\newblock
 doi:{\changeurlcolor{black}\href{https://doi.org/10.1007/978-3-030-32692-0_74}{\detokenize{10.1007/978-3-030-32692-0_74}}}.

\bibitem[Klein and Tourville(2012)]{Klein2012}
Klein, A.; Tourville, J.
\newblock 101 Labeled Brain Images and a Consistent Human Cortical Labeling
 Protocol.
\newblock {\em Front.   Neurosci.} {\bf 2012}, {\em 6}.
\newblock
 doi:{\changeurlcolor{black}\href{https://doi.org/10.3389/fnins.2012.00171}{\detokenize{10.3389/fnins.2012.00171}}}.

\bibitem[Lin et al.(2018)Lin, Yumer, Wang, Shechtman, and Lucey]{Lin2018}
Lin, C.H.; Yumer, E.; Wang, O.; Shechtman, E.; Lucey, S.
\newblock {ST}-{GAN}: Spatial Transformer Generative Adversarial Networks for
 Image Compositing.
\newblock In Proceedings of the IEEE Conference on Computer Vision and Pattern Recognition (CVPR), Salt Lake City, UT, USA, 18--22 June 2018.
\newblock
 doi:{\changeurlcolor{black}\href{https://doi.org/10.1109/cvpr.2018.00985}{\detokenize{10.1109/cvpr.2018.00985}}}.

\bibitem[Liu et al.(2015)Liu, Luo, Wang, and Tang]{Liu2015}
Liu, Z.; Luo, P.; Wang, X.; Tang, X.
\newblock Deep Learning Face Attributes in the Wild.
\newblock In Proceedings of the IEEE International Conference on Computer Vision (ICCV), Santiago, Chile, 11--18 December 2015.
\newblock
 doi:{\changeurlcolor{black}\href{https://doi.org/10.1109/iccv.2015.425}{\detokenize{10.1109/iccv.2015.425}}}.

\bibitem[Song et al.(2017)Song, Yu, Zeng, Chang, Savva, and
 Funkhouser]{Song2017}
Song, S.; Yu, F.; Zeng, A.; Chang, A.X.; Savva, M.; Funkhouser, T.
\newblock Semantic Scene Completion from a Single Depth Image.
\newblock In Proceedings of the IEEE Conference on Computer Vision and Pattern Recognition (CVPR), Honolulu, HI, USA, 21--26 July 2017.
\newblock
 doi:{\changeurlcolor{black}\href{https://doi.org/10.1109/cvpr.2017.28}{\detokenize{10.1109/cvpr.2017.28}}}.

\bibitem[Litany et al.(2018)Litany, Bronstein, Bronstein, and
 Makadia]{Litany2018}
Litany, O.; Bronstein, A.; Bronstein, M.; Makadia, A.
\newblock Deformable Shape Completion with Graph Convolutional Autoencoders.
\newblock In Proceedings of the IEEE Conference on Computer Vision and Pattern Recognition (CVPR), Salt Lake City, UT, USA, 18--22 June 2018.
\newblock
 doi:{\changeurlcolor{black}\href{https://doi.org/10.1109/cvpr.2018.00202}{\detokenize{10.1109/cvpr.2018.00202}}}.





\bibitem[Bogo et al.(2017)Bogo, Romero, Pons-Moll, and Black]{Bogo2017}
Bogo, F.; Romero, J.; Pons-Moll, G.; Black, M.J.
\newblock Dynamic {FAUST}: Registering Human Bodies in Motion.
\newblock In Proceedings of the IEEE Conference on Computer Vision and Pattern Recognition (CVPR), Honolulu, HI, USA, 21--26 July 2017.
\newblock
 doi:{\changeurlcolor{black}\href{https://doi.org/10.1109/cvpr.2017.591}{\detokenize{10.1109/cvpr.2017.591}}}.

\bibitem[Liu et al.(2019)Liu, Qi, and Guibas]{Liu2018}
Liu, X.; Qi, C.R.; Guibas, L.J.
\newblock {FlowNet}3D: Learning Scene Flow in 3D Point Clouds.
\newblock In Proceedings of the IEEE/CVF Conference on Computer Vision and Pattern Recognition (CVPR), Long Beach, CA, USA, 15--20 June 2019.
\newblock
 doi:{\changeurlcolor{black}\href{https://doi.org/10.1109/cvpr.2019.00062}{\detokenize{10.1109/cvpr.2019.00062}}}.

\bibitem[Mayer et al.(2016)Mayer, Ilg, Hausser, Fischer, Cremers,
 Dosovitskiy, and Brox]{Mayer2016}
Mayer, N.; Ilg, E.; Hausser, P.; Fischer, P.; Cremers, D.; Dosovitskiy, A.;
 Brox, T.
\newblock A Large Dataset to Train Convolutional Networks for Disparity,
 Optical Flow, and Scene Flow Estimation.
\newblock In Proceedings of the IEEE Conference on Computer Vision and Pattern Recognition (CVPR), Las Vegas, NV, USA, 27--30 June 2016.
\newblock
 doi:{\changeurlcolor{black}\href{https://doi.org/10.1109/cvpr.2016.438}{\detokenize{10.1109/cvpr.2016.438}}}.

\bibitem[Mahapatra et al.(2018)Mahapatra, Antony, Sedai, and
 Garnavi]{Mahapatra2018}
Mahapatra, D.; Antony, B.; Sedai, S.; Garnavi, R.
\newblock Deformable medical image registration using generative adversarial
 networks.
\newblock In Proceedings of the 2018 IEEE 15th International Symposium on Biomedical Imaging (ISBI 2018), Washington, DC, USA, 4--7 April 2018.   
\newblock
 doi:{\changeurlcolor{black}\href{https://doi.org/10.1109/isbi.2018.8363845}{\detokenize{10.1109/isbi.2018.8363845}}}.

\bibitem[Alipour et al.(2012)Alipour, Rabbani, and
 Akhlaghi]{HajebMohammadAlipour2012}
Alipour, S.H.M.; Rabbani, H.; Akhlaghi, M.R.
\newblock Diabetic Retinopathy Grading by Digital Curvelet Transform.
\newblock {\em Comput.   Math.   Methods Med.} {\bf 2012}, {\em 2012}, 1--11.
\newblock
 doi:{\changeurlcolor{black}\href{https://doi.org/10.1155/2012/761901}{\detokenize{10.1155/2012/761901}}}.

\bibitem[Radau et al.(2009)Radau, Lu, Connelly, Paul, Dick, and
 Wright]{radau2009evaluation}
Radau, P.; Lu, Y.; Connelly, K.; Paul, G.; Dick, A.; Wright, G.
\newblock Evaluation framework for algorithms segmenting short axis cardiac
 MRI.
\newblock {\em MIDAS J.   Cardiac MR Left Ventricle Segm.   Chall.} {\bf
 2009}, {\em 49}.   

\bibitem[Ofir et al.(2018)Ofir, Silberstein, Levi, Rozenbaum, Keller, and
 Bar]{Ofir2018}
Ofir, N.; Silberstein, S.; Levi, H.; Rozenbaum, D.; Keller, Y.; Bar, S.D.
\newblock Deep Multi-Spectral Registration Using Invariant Descriptor Learning.
\newblock In Proceedings of the 2018 25th IEEE International Conference on Image Processing (ICIP), Athens, Greece, 7--10 October 2018 
\newblock
 doi:{\changeurlcolor{black}\href{https://doi.org/10.1109/icip.2018.8451640}{\detokenize{10.1109/icip.2018.8451640}}}.

\bibitem[Krizhevsky \em{et~al.}(2009)Krizhevsky, Hinton,
et~al.]{krizhevsky2009learning}
Krizhevsky, A.; Hinton, G.; others.
\newblock Learning multiple layers of features from tiny images.
\newblock Technical report,  2009.

\bibitem[Brown and Susstrunk(2011)]{Brown2011}
Brown, M.; Susstrunk, S.
\newblock Multi-spectral {SIFT} for scene category recognition.
\newblock In Proceedings of the {CVPR}, Providence, RI, USA, 20--25 June 2011.
\newblock
 doi:{\changeurlcolor{black}\href{https://doi.org/10.1109/cvpr.2011.5995637}{\detokenize{10.1109/cvpr.2011.5995637}}}.

\bibitem[Yan et al.(2018)Yan, Xu, Rastinehad, and Wood]{Yan2018}
Yan, P.; Xu, S.; Rastinehad, A.R.; Wood, B.J.
\newblock Adversarial Image Registration with Application for {MR} and {TRUS}
 Image Fusion.   In {\em MLMI}; Springer International Publishing: Cham, Switzerland, 2018; pp.   197--204.
\newblock
 doi:{\changeurlcolor{black}\href{https://doi.org/10.1007/978-3-030-00919-9_23}{\detokenize{10.1007/978-3-030-00919-9_23}}}.

\bibitem[Wang et al.(2018)Wang, Rosa, Yang, Wang, Trigoni, and
 Markham]{Wang2018b}
Wang, Z.; Rosa, S.; Yang, B.; Wang, S.; Trigoni, N.; Markham, A.
\newblock 3D-{PhysNet}: Learning the Intuitive Physics of Non-Rigid Object
 Deformations.
\newblock \emph{arXiv} \textbf{2018}, arXiv:1805.00328.

\bibitem[Aoki et al.(2019)Aoki, Goforth, Srivatsan, and Lucey]{Aoki2019}
Aoki, Y.; Goforth, H.; Srivatsan, R.A.; Lucey, S.
\newblock {PointNetLK}: Robust {\&} Efficient Point Cloud Registration Using
 {PointNet}.
\newblock In Proceedings of the IEEE/CVF Conference on Computer Vision and Pattern Recognition (CVPR), Long Beach, CA, USA, 15--20 June 2019.
\newblock
 doi:{\changeurlcolor{black}\href{https://doi.org/10.1109/cvpr.2019.00733}{\detokenize{10.1109/cvpr.2019.00733}}}.

\bibitem[Chang and Pham(2019)]{Chang2019}
Chang, W.C.; Pham, V.T.
\newblock 3-D Point Cloud Registration Using Convolutional Neural Networks.
\newblock {\em Appl.   Sci.} {\bf 2019}, {\em 9}, 3273.
\newblock
 doi:{\changeurlcolor{black}\href{https://doi.org/10.3390/app9163273}{\detokenize{10.3390/app9163273}}}.

\bibitem[Guan et al.(2019)Guan, Meng, Xie, Wang, Sun, and Wang]{Guan2019}
Guan, S.; Meng, C.; Xie, Y.; Wang, Q.; Sun, K.; Wang, T.
\newblock Deformable Cardiovascular Image Registration via Multi-Channel
 Convolutional Neural Network.
\newblock {\em {IEEE} Access} {\bf 2019}, {\em 7}, 17524--17534.
\newblock
 doi:{\changeurlcolor{black}\href{https://doi.org/10.1109/access.2019.2894943}{\detokenize{10.1109/access.2019.2894943}}}.

\bibitem[Jack et al.(2019)Jack, Pontes, Sridharan, Fookes, Shirazi, Maire,
 and Eriksson]{Jack2019}
Jack, D.; Pontes, J.K.; Sridharan, S.; Fookes, C.; Shirazi, S.; Maire, F.;
 Eriksson, A.
\newblock Learning Free-Form Deformations for 3D Object Reconstruction.   In {\em
 {ACCV}}; Springer: Berlin/Heidelberg, Germany, 2019; pp.   317--333.
\newblock
 doi:{\changeurlcolor{black}\href{https://doi.org/10.1007/978-3-030-20890-5_21}{\detokenize{10.1007/978-3-030-20890-5_21}}}.

\bibitem[Deng et al.(2009)Deng, Dong, Socher, Li, Li, and
 Fei-Fei]{Deng2009}
Deng, J.; Dong, W.; Socher, R.; Li, L.J.; Li, K.; Fei-Fei, L.
\newblock {ImageNet}: A large-scale hierarchical image database.
\newblock In Proceedings of the 2009 IEEE Conference on Computer Vision and Pattern Recognition, Miami, FL, USA, 20--25 June 2009.   
\newblock
 doi:{\changeurlcolor{black}\href{https://doi.org/10.1109/cvpr.2009.5206848}{\detokenize{10.1109/cvpr.2009.5206848}}}.

\bibitem[Schaffert et al.(2019)Schaffert, Wang, Fischer, Borsdorf, and
 Maier]{Schaffert2019}
Schaffert, R.; Wang, J.; Fischer, P.; Borsdorf, A.; Maier, A.
\newblock Metric-Driven Learning of Correspondence Weighting for 2-D/3-D Image
 Registration.   In {\em Lect.   Notes Comput.   Sci.}; Springer International Publishing: Cham, Switzerland, 2019;
 pp.   140--152.
\newblock
 doi:{\changeurlcolor{black}\href{https://doi.org/10.1007/978-3-030-12939-2_11}{\detokenize{10.1007/978-3-030-12939-2_11}}}.

\bibitem[Smirnov et al.(2019)Smirnov, Bessmeltsev, and
 Solomon]{Smirnov2019}
Smirnov, D.; Bessmeltsev, M.; Solomon, J.
\newblock Deep Sketch-Based Modeling of Man-Made Shapes.
\newblock {\em arXiv} {\bf 2019}, arXiv:1906.12337.

\bibitem[Yang et al.(2019)Yang, Huang, Hao, Liu, Belongie, and
 Hariharan]{Yang2019}
Yang, G.; Huang, X.; Hao, Z.; Liu, M.Y.; Belongie, S.; Hariharan, B.
\newblock Pointflow: 3d point cloud generation with continuous normalizing
 flows.
\newblock In Proceedings of the IEEE/CVF International Conference on Computer Vision (ICCV), Seoul, Korea, 27--28 October 2019; pp.   4541--4550.

\bibitem[Wang and Solomon(2019)]{Wang_2019_ICCV}
Wang, Y.; Solomon, J.M.
\newblock Deep Closest Point: Learning Representations for Point Cloud
 Registration.
\newblock In Proceedings of the IEEE/CVF International Conference on Computer Vision (ICCV), Seoul, Korea, 27--28 October 2019.

\bibitem[Wang et al.(2019)Wang, Ceylan, Mech, and Neumann]{wang20193dn}
Wang, W.; Ceylan, D.; Mech, R.; Neumann, U.
\newblock 3DN: 3D Deformation Network.
\newblock In Proceedings of the IEEE/CVF Conference on Computer Vision and Pattern Recognition (CVPR), Long Beach, CA, USA, 15--20 June 2019.
\newblock
 doi:{\changeurlcolor{black}\href{https://doi.org/10.1109/cvpr.2019.00113}{\detokenize{10.1109/cvpr.2019.00113}}}.

\bibitem[Wang and Fang(2019)]{wang2019coherent}
Wang, L.; Fang, Y.
\newblock Coherent point drift networks: Unsupervised learning of non-rigid
 point set registration.
\newblock {\em arXiv} {\bf 2019}, arXiv:1906.03039.

\bibitem[Pais et al.(2020)Pais, Ramalingam, Govindu, Nascimento,
 Chellappa, and Miraldo]{Pais2019}
Pais, G.D.; Ramalingam, S.; Govindu, V.M.; Nascimento, J.C.; Chellappa, R.;
 Miraldo, P.
\newblock 3DRegNet: A Deep Neural Network for 3D Point Registration.
\newblock In Proceedings of the IEEE/CVF Conference on Computer Vision and Pattern Recognition (CVPR), Seattle, WA, USA, 14--19 June 2020; pp.   7193--7203.

\bibitem[Choi et al.(2015)Choi, Zhou, and Koltun]{SungjoonChoi2015}
Choi, S.; Zhou, Q.Y.; Koltun, V.
\newblock Robust reconstruction of indoor scenes.
\newblock In Proceedings of the IEEE Conference on Computer Vision and Pattern Recognition (CVPR), Boston, MA, USA, 7--12 June 2015.
\newblock
 doi:{\changeurlcolor{black}\href{https://doi.org/10.1109/cvpr.2015.7299195}{\detokenize{10.1109/cvpr.2015.7299195}}}.

\bibitem[Zhou et al.(2014)Zhou, Lapedriza, Xiao, Torralba, and
 Oliva]{zhou2014learning}
Zhou, B.; Lapedriza, A.; Xiao, J.; Torralba, A.; Oliva, A.
\newblock Learning deep features for scene recognition using places database.
\newblock In Proceedings of the Advances in Neural Information Processing Systems 27 (NIPS 2014), {Montreal, MTL, Canada, }8--13 December 2014; pp.   487--495.   

\bibitem[Li et al.(2020)Li, Pontes, and Lucey]{li2020deterministic}
Li, X.; Pontes, J.K.; Lucey, S.
\newblock Deterministic PointNetLK for Generalized Registration.
\newblock {\em arXiv} {\bf 2020}, arXiv:2008.09527.

\bibitem[Yuan et al.(2020)Yuan, Eckart, Kim, Jampani, Fox, and
 Kautz]{yuan2020deepgmr}
Yuan, W.; Eckart, B.; Kim, K.; Jampani, V.; Fox, D.; Kautz, J.
\newblock DeepGMR: Learning Latent Gaussian Mixture Models for Registration.
\newblock {\em arXiv} {\bf 2020}, arXiv:2008.09088.

\bibitem[Handa et al.(2014)Handa, Whelan, McDonald, and
 Davison]{handa2014benchmark}
Handa, A.; Whelan, T.; McDonald, J.; Davison, A.J.
\newblock A benchmark for RGB-D visual odometry, 3D reconstruction and SLAM.
\newblock In Proceedings of the 2014 IEEE International Conference on Robotics and Automation (ICRA), Hong Kong, China, 31 May--7 June 2014; pp.   1524--1531.

\bibitem[Zhang et al.(2020)Zhang, Jian, Chen, and Yang]{zhang2020deform}
Zhang, X.; Jian, W.; Chen, Y.; Yang, S.
\newblock Deform-GAN: An Unsupervised Learning Model for Deformable
 Registration.
\newblock {\em arXiv} {\bf 2020}, arXiv:2002.11430.

\bibitem[Menze et al.(2015)Menze, Jakab, Bauer, Kalpathy-Cramer, Farahani,
 Kirby, Burren, Porz, Slotboom, Wiest, et al.]{menze2015multimodal}
Menze, B.H.; Jakab, A.; Bauer, S.; Kalpathy-Cramer, J.; Farahani, K.; Kirby,
 J.; Burren, Y.; Porz, N.; Slotboom, J.; Wiest, R.; others.
\newblock The multimodal brain tumor image segmentation benchmark (BRATS).
\newblock {\em IEEE Trans.   Med.   Imaging} {\bf 2015}, {\em
 34}, 1993--2024.
 
 \bibitem[Saval-Calvo et al.(2015)Saval-Calvo, Orts-Escolano, Azorin-Lopez,
 Garcia-Rodriguez, Fuster-Guillo, Morell-Gimenez, and Cazorla]{SavalCalvo2015}
Saval-Calvo, M.; Orts-Escolano, S.; Azorin-Lopez, J.; Garcia-Rodriguez, J.;
 Fuster-Guillo, A.; Morell-Gimenez, V.; Cazorla, M.
\newblock Non-rigid point set registration using color and data downsampling.
\newblock In Proceedings of the 2015 International Joint Conference on Neural Networks (IJCNN), Killarney, Ireland, 12--17 July 2015.
\newblock
 doi:{\changeurlcolor{black}\href{https://doi.org/10.1109/ijcnn.2015.7280765}{\detokenize{10.1109/ijcnn.2015.7280765}}}.

\bibitem[LeCun et al.(2015)LeCun, Bengio, and Hinton]{LeCun2015}
LeCun, Y.; Bengio, Y.; Hinton, G.
\newblock Deep learning.
\newblock {\em Nature} {\bf 2015}, {\em 521}, 436--444.
\newblock
 doi:{\changeurlcolor{black}\href{https://doi.org/10.1038/nature14539}{\detokenize{10.1038/nature14539}}}.

\bibitem[Alom et al.(2018)Alom, Taha, Yakopcic, Westberg, Sidike, Nasrin,
 Van Esesn, Awwal, and Asari]{Alom2018}
Alom, M.Z.; Taha, T.M.; Yakopcic, C.; Westberg, S.; Sidike, P.; Nasrin, M.S.;
 Van Esesn, B.C.; Awwal, A.A.S.; Asari, V.K.
\newblock The history began from alexnet: A comprehensive survey on deep
 learning approaches.
\newblock {\em arXiv} {\bf 2018}, arXiv:1803.01164.

\bibitem[Bench-Capon(1990)]{Bench-Capon1990}
Bench-Capon, T.J.M.
\newblock {Knowledge representation: an approach to artificial intelligence}.
\newblock {\em { A.P.I.C.   Series}} {\bf 1990}, {\em 32}, 220.   

\bibitem[Norman(1987)]{norman1983}
Norman, D.A.   Some Observations on Mental Models.
\newblock In {\em Human-Computer Interaction: A Multidisciplinary Approach};
 Morgan Kaufmann Publishers Inc.: San Francisco, CA, USA, 1987; pp.   241–244.

\bibitem[Nersessian(1992)]{nersessian1992scientists}
Nersessian, N.
\newblock How do scientists think?Capturing the dynamics of conceptual change
 in science.
\newblock {\em \mbox{Cogn.   Model.   Sci.}} {\bf 1992}, {\em 15}, 3--44.

\bibitem[Greca and Moreira(2000)]{Greca2000}
Greca, I.M.; Moreira, M.A.
\newblock Mental models, conceptual models, and modelling.
\newblock {\em Int.   J.   Sci.   Educ.} {\bf 2000}, {\em 22}, 1--11.
\newblock
 doi:{\changeurlcolor{black}\href{https://doi.org/10.1080/095006900289976}{\detokenize{10.1080/095006900289976}}}.

\bibitem[Mahadevan(2018)]{mahadevan2018imagination}
Mahadevan, S.
\newblock Imagination machines: A new challenge for artificial intelligence.
\newblock In Proceedings of the AAAI, New Orleans, LA, USA, 2--7 February 2018.

\bibitem[Qi et al.(2017)Qi, Yi, Su, and Guibas]{Qi2017a}
Qi, C.R.; Yi, L.; Su, H.; Guibas, L.J.
\newblock Pointnet++:Deep hierarchical feature learning on point sets in a
 metric space.
\newblock In Proceedings of the Advances in Neural Information Processing Systems, {Long Beach, CA, USA}, 4--9 December 2017; pp.   5099--5108.   

\bibitem[Charles et al.(2017)Charles, Su, Kaichun, and
 Guibas]{Charles2017}
Charles, R.Q.; Su, H.; Kaichun, M.; Guibas, L.J.
\newblock {PointNet}: Deep Learning on Point Sets for 3D Classification and
 Segmentation.
\newblock In Proceedings of the IEEE Conference on Computer Vision and Pattern Recognition (CVPR), Honolulu, HI, USA, 21--26 July 2017.
\newblock
 doi:{\changeurlcolor{black}\href{https://doi.org/10.1109/cvpr.2017.16}{\detokenize{10.1109/cvpr.2017.16}}}.

\bibitem[Maturana and Scherer(2015)]{Maturana2015}
Maturana, D.; Scherer, S.
\newblock {VoxNet}: A 3D Convolutional Neural Network for real-time object
 recognition.
\newblock In Proceedings of the 2015 IEEE/RSJ International Conference on Intelligent Robots and Systems (IROS), Hamburg, Germany, 28 September--2 October 2015.   
\newblock
 doi:{\changeurlcolor{black}\href{https://doi.org/10.1109/iros.2015.7353481}{\detokenize{10.1109/iros.2015.7353481}}}.

\bibitem[Xu et al.(2018)Xu, Fan, Xu, Zeng, and Qiao]{Xu2018}
Xu, Y.; Fan, T.; Xu, M.; Zeng, L.; Qiao, Y.
\newblock {SpiderCNN}: Deep Learning on Point Sets with Parameterized
 Convolutional Filters.   In {\em {ECCV}}; Springer: Berlin/Heidelberg, Germany, 2018; pp.   90--105.
\newblock
 doi:{\changeurlcolor{black}\href{https://doi.org/10.1007/978-3-030-01237-3_6}{\detokenize{10.1007/978-3-030-01237-3_6}}}.

\bibitem[Lee et al.(2019)Lee, Kim, and Kim]{Lee2019}
Lee, S.H.; Kim, H.U.; Kim, C.S.
\newblock {ELF}-Nets: Deep Learning on Point Clouds Using Extended Laplacian
 Filter.
\newblock {\em {IEEE} Access} {\bf 2019}, {\em 7}, 156569--156581.
\newblock
 doi:{\changeurlcolor{black}\href{https://doi.org/10.1109/access.2019.2949785}{\detokenize{10.1109/access.2019.2949785}}}.

\bibitem[He et al.(2016)He, Zhang, Ren, and Sun]{He2016}
He, K.; Zhang, X.; Ren, S.; Sun, J.
\newblock Deep Residual Learning for Image Recognition.
\newblock In Proceedings of the IEEE Conference on Computer Vision and Pattern Recognition (CVPR), Las Vegas, NV, USA, 27--30 June 2016.
\newblock
 doi:{\changeurlcolor{black}\href{https://doi.org/10.1109/cvpr.2016.90}{\detokenize{10.1109/cvpr.2016.90}}}.

\bibitem[Jaderberg et al.(2015)Jaderberg, Simonyan, Zisserman,
 et al.]{jaderberg2015spatial}
Jaderberg, M.; Simonyan, K.; Zisserman, A.
\newblock Spatial transformer networks.
\newblock In Proceedings of the Advances in Neural Information Processing Systems,{ Montreal, MTL, Canada, }7--12 December 2015; pp.   2017--2025.   

\bibitem[Shan et al.(2017)Shan, Yan, Guo, Chang, Fan, Xu,
 et al.]{shan2017unsupervised}
Shan, S.; Yan, W.; Guo, X.; Chang, E.I.; Fan, Y.; Xu, Y.
\newblock Unsupervised end-to-end learning for deformable medical image
 registration.
\newblock {\em arXiv} {\bf 2017}, arXiv:1711.08608.

\bibitem[Balakrishnan et al.(2018)Balakrishnan, Zhao, Sabuncu, Dalca, and
 Guttag]{Balakrishnan2018}
Balakrishnan, G.; Zhao, A.; Sabuncu, M.R.; Dalca, A.V.; Guttag, J.
\newblock An Unsupervised Learning Model for Deformable Medical Image
 Registration.
\newblock In Proceedings of the IEEE Conference on Computer Vision and Pattern Recognition (CVPR), Salt Lake City, UT, USA, 18--22 June 2018.
\newblock
 doi:{\changeurlcolor{black}\href{https://doi.org/10.1109/cvpr.2018.00964}{\detokenize{10.1109/cvpr.2018.00964}}}.

\bibitem[Ronneberger et al.(2015)Ronneberger, Fischer, and
 Brox]{Ronneberger2015}
Ronneberger, O.; Fischer, P.; Brox, T.
\newblock U-Net:Convolutional Networks for Biomedical Image Segmentation.   In
 {\em Lect.   Notes Comput.   Sci.}; Springer International Publishing: Cham, Switzerland, 2015; pp.   234--241.
\newblock
 doi:{\changeurlcolor{black}\href{https://doi.org/10.1007/978-3-319-24574-4_28}{\detokenize{10.1007/978-3-319-24574-4_28}}}.
 

\bibitem[Howard et al.(2017)Howard, Zhu, Chen, Kalenichenko, Wang, Weyand,
 Andreetto, and Adam]{howard2017mobilenets}
Howard, A.G.; Zhu, M.; Chen, B.; Kalenichenko, D.; Wang, W.; Weyand, T.;
 Andreetto, M.; Adam, H.
\newblock Mobilenets: Efficient conv.   neural networks for mobile vision
 applications.
\newblock {\em arXiv} {\bf 2017}, arXiv:1704.04861.

\bibitem[Lucas and Kanade(1981)]{lucas1981iterative}
Lucas, B.D.; Kanade, T.
\newblock An Iterative Image Registration Technique with an Application to
 Stereo Vision.
\newblock In \emph{IJCAI}; Morgan Kaufmann Publishers Inc.: San Francisco, CA, USA,
 1981; IJCAI’81, pp.   674--679.

\bibitem[Wang et al.(2019)Wang, Sun, Liu, Sarma, Bronstein, and
 Solomon]{Wang2019}
Wang, Y.; Sun, Y.; Liu, Z.; Sarma, S.E.; Bronstein, M.M.; Solomon, J.M.
\newblock Dynamic Graph {CNN} for Learning on Point Clouds.
\newblock {\em {ACM} Trans.   Graph.} {\bf 2019}, {\em 38}, 1--12.
\newblock
 doi:{\changeurlcolor{black}\href{https://doi.org/10.1145/3326362}{\detokenize{10.1145/3326362}}}.
 
\bibitem[Lu and Shi(2020)]{lu2020deep}
 Lu, H.; Shi, H.
\newblock Deep Learning for 3D Point Cloud Understanding: A Survey.
\newblock {\em arXiv preprint arXiv:2009.08920} {\bf 2020}.

\end{thebibliography}

\reftitle{References}

\end{document}